\documentclass[sigconf, nonacm]{acmart}
\AtBeginDocument{%
  }

\setcopyright{acmlicensed}
\copyrightyear{2018}
\acmYear{2018}
\acmDOI{XXXXXXX.XXXXXXX}
\acmConference[SETN '26]{14th EETN Conference on Artificial Intelligence}{September 09--11, 2026}{Chania, Greece}
\acmISBN{978-1-4503-XXXX-X/2018/06}

\begin{document}

\title{Human-Centric Reflective Architecture for Human-AI  Collaborative  Decision-Making}


\author{Andreas Kouridakis}
\email{andreaskouridakis@outlook.com}
\affiliation{
  \institution{University of Piraeus}
  \city{Piraeus}
  \country{Greece}}

\author{Dimitrios Patiniotis Spyropoulos}
\email{dimpatspy@gmail.com}
\affiliation{
  \institution{University of Piraeus}
  \city{Piraeus}
  \country{Greece}}
  
\author{George A. Vouros}
\email{georgev@unipi.gr}
\orcid{0000-0001-5451-622X}
\affiliation{
  \institution{University of Piraeus}
  \city{Piraeus}
  \country{Greece}}
%\renewcommand{\shortauthors}{Kouridakis et al.}

\begin{abstract}
  The use of Large Language Models (LLMs) across diverse areas of human activity—ranging from everyday tasks to safety-critical applications—aims to enhance decision-making effectiveness with minimal human feedback. Concurrently, it seeks to align decisions with human expectations, preferences, and needs while mitigating risks associated with AI non-determinism. However, humans frequently over- or under-rely on AI recommendations, and current AI systems remain poorly calibrated to human expectations. To address these challenges, we introduce a human-AI collaborative decision-making framework designed to augment human capabilities and align AI agents with human preferences and expectations. Specifically, this paper (a) formulates the collaborative decision-making task as a stochastic game between an AI agent and a human player, and (b) proposes the Human-Centric Reflective Architecture (HCRA), which integrates human-calibrated models with reinforcement learning agents that leverage linguistic feedback in an iterative, reflective process. Evaluation results demonstrate that HCRA significantly enhances decision-making effectiveness and delivers high-quality recommendations.
\end{abstract}




\received{20 February 2007}
\received[revised]{12 March 2009}
\received[accepted]{5 June 2009}

\maketitle

\section{Introduction}
In many collaborative decision-making environments, humans and AI agents engage in an interactive feedback loop. This iterative process aims to: (a) guide agents toward human-acceptable recommendations, (b) enable users to express and refine their preferences and expectations, and (c) allow agents to learn from historical interactions. Such tight interaction is vital in high-stakes domains (e.g., \cite{10.1145/3313831.3376718}, \cite{FISCHERABAIGAR2024101976}) where humans must retain ultimate decision-making authority. 

However, effective human-AI collaboration remains a challenge. When partnering with AI, human performance often falls short of expectations \cite{10.1145/3479552}. While AI explanations can influence user decisions \cite{10.5555/3737916.3738079}, they frequently fail to improve system understanding or properly calibrate trust \cite{10.1145/3479552} \cite{10.1145/3377325.3377498} \cite{10.1145/3411764.3445717} \cite{10.1145/3351095.3372852} \cite{10.1145/3397481.3450650} \cite{DBLP:journals/corr/abs-2212-06823}. This shortfall is especially critical when users must dedicate limited cognitive bandwidth to time-sensitive, high-consequence problems \cite{PAPADOPOULOS2024121234}. Furthermore, high recommendation accuracy alone is insufficient to build trust; effective collaboration requires AI systems to optimize for human utility alongside core task objectives \cite{Bansal_Nushi_Kamar_Horvitz_Weld_2021}, \cite{10.5555/3600270.3600559}..

Focusing on LLM-assisted agents, our work is motivated by two core limitations. First, miscalibrated human expectations regarding AI capabilities often trigger over- or under-reliance \cite{Bansal_Nushi_Kamar_Lasecki_Weld_Horvitz_2019} \cite{zhang2023taking} \cite{10.1145/3706598.3714097}. Second, rather than relying solely on large-scale training followed by aggregated alignment (e.g., RLHF \cite{ziegler2020finetuninglanguagemodelshuman} \cite{10.5555/3600270.3602281}), we leverage test-time scaling \cite{snell2024scalingllmtesttimecompute} \cite{openai2024openaio1card} and contextual test-time tuning. This approach establishes a human-centric workflow where high-quality decisions are made in-context based on user utility, enabling agents to dynamically refine recommendation quality and mitigate issues stemming from LLM non-determinism.

To address these issues, we propose a human-centric reflective architecture (HCRA) for collaborative decision-making that integrates reinforcement learning (RL) language agents with human-calibrated models \cite{10.5555/3600270.3600559}, in an iterative reflective process \cite{xu2025largereasoningmodelssurvey}\footnote{Datasets and code are available in \url{https://github.com/AILabDsUnipi/HCRA}}.

This article makes the following contributions:

- Game-Theoretic Formulation: We model the collaborative decision-making process as a stochastic game between an AI agent and a human. Within this framework, we define the human-utility optimized for user expectations and preferences. This utility maximization criterion provides formal convergence and termination guarantees for the iterative decision-making process.

- Architecture Design: We propose HCRA, a framework that optimizes human-AI collaboration through an iterative reflection process. HCRA tunes AI recommendations at test time by leveraging historical interactions, an AI recommendation calibration model, and a human acceptance model.

- Empirical Evaluation: We validate HCRA within the domain of tourism recommendations—a setting where inaccuracies can have real-world consequences.\footnote{See, for example, \url{https://www.webpronews.com/ai-hallucinations-in-travel-apps-lead-to-fake-landmarks-and-dangers/}.} Our results demonstrate the framework's effectiveness and recommendation quality, highlighting the vital role of human behavior modeling and human-centric objective functions.

%%%%%%%%%%%%%%%%%%%%%%%%%%%%%%%%%%%%%%%%%%%%%%%%%%%%%%%%%%%%%%%%%%%%%%%%

\section{Related work}
This work contributes to the growing and important efforts to assist AI-assisted decision-making \cite{10.1145/3593013.3594087}. The most basic paradigm here involves the AI agent to perform an assistive role by providing a recommendation that a human may accept or reject.

A critical challenge is whether  we can achieve human-AI collaboration that can  outperform the  human and AI alone,  contributing to high-quality recommendations via an effective decision-making process. Addressing this challenge, several recent studies propose different designs to help humans allocate appropriate trust to AI based on confidence scores. Empirical results suggest that people have poorly-calibrated self confidence  and cannot reliably reflect well-calibrated confidence scores. This motivates research questions in  \cite{Bansal_Nushi_Kamar_Horvitz_Weld_2021}, \cite{10.1145/3544548.3581058} and \cite{10.5555/3600270.3600559}, emphasizing on the importance of AI advice confidence for human decisions. This research is in contrast to shaping humans' self confidence for AI assisted decision-making (e.g. as in \cite{takayanagi2025impactfeasibilityselfconfidenceshaping}). 
The authors in \cite{10.1145/3544548.3581058} try to answer this question by proposing a framework that considers well-calibrated confidence scores from both humans and AI. However, the framework proposed in \cite{10.5555/3600270.3600559} highlights that modifying the confidence of an AI advice by exploiting a human behavior model can enhance  human-AI collaboration:  Authors propose  optimizing the transformation of the AI advice confidence so that the transformed confidence score matches  the human perception of AI confidence scores. In so doing they optimize  an AI system with respect to the human user, towards \textit{human-calibrated AI}. 

In addition to these efforts, authors in \cite{benz2024humanalignedcalibrationaiassisteddecision}, following a more fundamental approach, show that if the confidence values of (human) decision makers satisfy a natural alignment property with respect to the confidence they have on their own predictions (\textit{human-alignment}) then the decision makers can make optimal decisions.

Building on these results, in this paper we optimize  an AI language agent with respect to human behavior models. In contrast to \cite{10.5555/3600270.3600559} we consider that the human does not have any information on what the true advise could be (and, there is not necessarily an initial guess), but he/she may have certain requirements that should be satisfied by the final decision. Furthermore, in contrast to a single-step decision-making task, we formulate and implement an iterative reflective collaborative process which transparently, without human intervention, aims at  maximizing the utility of humans.

The reflective process is inspired by Reflexion \cite{10.5555/3666122.3666499}: It represents a novel paradigm in RL for language agents, with no requirements regarding gradient-based optimization methods. Reflexion enables agents to learn from trial-and-error through linguistic feedback rather than parameter updates,  without requiring expensive model fine-tuning. 

Reflexion is formalized as an iterative optimization process. As described in \cite{10.5555/3666122.3666499}, in any iteration of the reflective process $t$, an actor produces a trajectory $\tau_t$ to solve a sequential task, by interacting with the environment. An evaluator model produces a scalar reward signal $r_t$ that improves as task-specific performance increases. To amplify $r_t$ to a feedback form that can be used for improvement by an LLM, the self-reflection model analyzes the set of \{ ($\tau_i, r_i$), i=1,...,t\} to produce a summary $sr_t$, which is stored in a memory buffer. The actor, evaluator, and self-reflection models work together through iterations until the evaluator deems the final trajectory $\tau_T$ at iteration $T$,  to be correct. The memory components of Reflexion, i.e., the short term memory for trajectory history and the long term memory implementing  a persistent knowledge base that informs future decision-making,  are crucial to its effectiveness. 

%The framework consists of three core components: an Actor model that generates text-based recommendations and/or actions, an Evaluator that scores the outputs of the Actor, and a Self-Reflection model that produces verbal reinforcement cues. When an agent receives failure signals, the Self-Reflection component analyzes the Actor's trajectory and generates natural language summaries explaining mistakes and suggesting improvements. These reflections are stored in episodic memory and passed through the model's context window, creating a persistent knowledge base that informs future decision-making.

Reflexion's key innovation lies in transforming sparse rewards on trajectories into linguistic feedback signals. Reflexion's linguistic reflections provide specific insights into failure modes and suggest targeted corrective actions in an iterative process. %More specifically, when an agent receives a failure signal, it can infer that a specific action at a time step $t$ led to subsequent incorrect actions. The agent can then verbally state that it should have taken a different action at that step and store this experience in its memory. 
This iterative process of trial, error, self-reflection, and persistent memory, as shown in \cite{10.5555/3666122.3666499}, enables the agent to improve its decision-making ability and facilitates learning by addressing recommendation deficiencies.

Aiming to a human-centric decision-making process, we integrate well-trained predictive human models into the agent's reflective loop. These models steer the selection of recommendations toward human-centric objectives, aiming to maximize expected human utility, dictating the reflective iterative process termination conditions. Trained on empirical data gathered from real users, these models accurately simulate human acceptance behavior. By substituting simulated interactions for real-world responses, this approach allows the system to evaluate decisions without requiring continuous, manual human feedback at every step. Furthermore, we consider simple, one-shot decision-making tasks rather than sequential ones.

%%%%%%%%%%%%%%%%%%%%%%%%%%%%%%%%%%%%%%%%%%%%%%%%%%%%%%%%%%%%%%%%%%%%%%%%

\section{Problem formulation}
\label{sec:prob_form}

The iterative reflective decision-making process is formulated as a stochastic game between the AI and the human: 
At any iteration (time step) $t$, the process is at state $s_t$ and players choose actions as required by the decision-making process. They get their payoffs, and the process proceeds to the next iteration. 
Formally, the iterative reflective decision-making stochastic process has the following components:

-$S$: A set of games (states). $s_t \in S$ is specified to be of the form $<Qr_t, Cs, (Re_t,cf_t), g_t>$, where 
\begin{enumerate}
\item $Qr_t$ is the request at $t$, formed by the initial request $Qr_0$ specified by the human, and enhanced at subsequent iterations.

\item $Cs$ is a set of human preferences or constraints that should be 
satisfied by the agent recommendation. We may distinguish between preferences (a.k.a soft constraints) and hard constraints, although subsequently we refer to any of these as ``constraints". Constraints are expressed in natural language, but they can be formulated as ($attr$ $op$ $value$), where $attr$ is a real-world domain-specific attribute (e.g. time of task completion, duration, amount of resources, style/skill),  $op$ can be  any of the operators in $\{=, \leq, \geq, >, <\}$, or any domain-specific specification of a relation, and $value$ is any of the numerical or categorical values for $attr$ (e.g. $criminality$ $in$ $area$ $is$ $very$ $low$, $menu\_price<100$ Euros).

\item $(Re_t, cf_t)$ is the AI recommendation  at time step $t$, comprising the content $Re_t$ of the recommendation and the 
confidence $cf_t$ assessed by the agent. 

\item $g_t$ is the human-calibrated confidence of the recommendation, to be shown to the human. Actually, $g_t$ is a transformation of $cf_t$ to meet human-calibrated expectations from AI recommendations (explained subsequently).
\end{enumerate}
Constraints $Cs$ must be satisfied by the AI recommendation $Re_t$. In case there is not any violation, we refer to that situation as an  ``\textit{agreement}''. 

- $N$: This is the set of players, including the human (denoted by $h$), here represented by a human model,  and the AI  agent (denoted by $ag$).

 Although $ag$ has full access to the components of a state $s\in S$,  the human is assumed to observe the recommendation $Re_t$ and the human-calibrated confidence of the recommendation $g_t$, at any time step $t$.  %, although the human-calibrated confidence is affected by the assessed correctness of the recommendation. 

- Set of actions: $A=A^h \times A^{ag}$, where $A^i$ is a  set of actions available to player $i \in \{h,ag\}$. Specifically,  actions $a^h_t$ in $A^h$ are in $[0,1]$ and specify the probability of the human to accept the recommendation at that state. Given that humans are represented by a human behavior model, $a^h_t$  is the $predicted$ probability of the human to accept the recommendation at time step $t$ and is provided by the function $h_{acc}: S \rightarrow [0,1]$. Actions $a^{ag}_t$ are in $A^{ag}=\{Reflective\_text(s_t)| s_t \in S, t=0,1,2...\}$, which is the set of reflective texts that can be provided as linguistic feedback at any time step $t$. 

- $P:S \times A \times S \rightarrow [0,1]$ is the state transition function from state $s_t$ to state $s_{t+1}$ after the execution of the joint players' action $a_t=(a_t^h,a_t^{ag} ) \in A$.

- Reward functions  $r^h$ and  $r^{ag}$: $r^h: S  \rightarrow \mathbb{R}$ 
is a real-valued payoff function for the  human player, and $r^{ag}: S  \rightarrow <\mathbb{R} \times \mathbb{R}>$ is the payoff function for the  agent comprising per $s_t$ the correctness and the agreement $assessments$ with regards to $Re_t$, denoted by $\widehat{Corr_t}$ and $\widehat{Aggr_t}$, respectively. We consider these as assessments, given that no player has access to the ground truth.

Although alternative formulations of the stochastic process are possible (e.g. specifying agent actions to be the set of possible recommendations), the above specification emphasizes on the importance of the linguistic reflective feedback provided by RL language agents during the collaborative process.  

%\textbf{Later?} The action $a^{ag}_t$ of the AI agent (i.e. the reflective text) in a state $s_t$ depends on the following information: the reward $r^{ag}(s_t,a_{t})$ (i.e. correctness and agreement assessments),  the predicted probability $h_{acc}(s_t)$ for human acceptance at state $s_t$, the human-calibrated confidence of the recommendation $g_t$ shown to the human, and historical interactions. %It must be noted that since there is not access to the ground correctness and agreement of any AI recommendation, only an assessment on correctness and agreement can be made.
%As formalized above, it is assumed that the AI reward comprises an aggregated payoff value. 

\subsection{Rewards}

The reward $r^{ag}(s_t)$ that the AI agent gets after the execution of the joint action $a_t=(a^{ag}_{t-1},a^{h}_{t-1})$ in a state $s_{t-1}$ results from the evaluation of the  state $s_{t}$ in terms of correctness and agreement to constraints. This is done by an evaluation component, whose functionality is specified subsequently.

%As pointed out above,  $r{ag}$ comprises two real-value components: $r_{ag}(s_t,a_t)=<\widehat{Corr_t}, \widehat{Aggr_t}>$.

The reward of the human in iteration $t$ is defined as follows:
\[
    r^h(s_t)= c(s_t) \times log(h_{acc}(s_t)) +(1-c(s_t))\times log(1-h_{acc}(s_t))
\]
\noindent where,  $c(s_t) \in \{0,1\}$ is the $assessed$ class of the AI recommendation at  state $s_t$, depending on the assessed correctness and the assessed agreement of the AI recommendation $Re_t$.  We  consider $c(s_t)=1$ when the recommendation $Re_t$ is assessed to be correct and with agreement to constraints, and $c(s_t)=0$ in all other cases. In a more general case, we can consider $c(s_t)$  to be the probability that at state $s_t$,  $Re_t$ is correct and with agreement, but this is not the case in this work. 
This reward function assigns a high penalty when the human (model) either (a) accepts a recommendation that is assessed to be incorrect or with no agreement to constraints, or (b) does not accept a recommendation that is assessed to be correct and with  agreement to constraints. Humans get the highest reward when they accept a proposal that is assessed to be correct and in agreement to constraints.

It must be noted that state transitions and thus rewards players get depend on the joint action of players. Also, $r^h$ depends on the components of the agent reward $r^{ag}$.

\subsection{Objective}
The objective of the overall task is to maximize the human utility, or else, to minimize the expected  human loss $\mathcal{L}_T$, after $T$ iterations.
To formulate this, we can consider that at each time step $t$,  the human player plays the lottery
\[ [L_t: h_{acc}(s_t), \mathcal{L}_{refl}(s_t):(1-h_{acc}(s_t))] \]
 \noindent where, if the human accepts the decision at $t$ with probability $h_{acc}(s_t)$ the loss experienced by the human is $L(s_t)=-r^h(s_t)$. Else, with probability $(1-h_{acc}(s_t))$, the human experiences the expected loss $\mathcal{L}_{refl}(s_t)=\big(  h_{acc}(s_t)L(s_{t+1})+(1- h_{acc}(s_{t+1})\mathcal{L}_{refl}(s_{t+1}) ) \big)$ by proceeding to subsequent iterations of the process. 
In so doing, at the final iteration $T$ the expected human loss is
\begin{align*}
\mathcal{L}_T=
    &[h_{acc}(s_T)\prod^{T-1}_{j=1}(1-h_{acc}(s_j)) L(s_T)]+\\&[\sum^{T-1}_{i=1}h_{acc}(s_i)L(s_i)\prod_{j-1}^{i-1}(1-h_{acc}(s_j)] 
\end{align*}
The first term in brackets specifies the expected loss in case the human accepts the recommendation $Re_T$ generated at the final time step $T$ (previous recommendations have not been accepted). The second bracket specifies the expected loss of accepting any of the previous  recommendations generated at time steps $1,...,(T-1)$.

 Ideally, the final iteration $T$  occurs  when the recommendation at  $T$ is accepted with probability equal to 1. However, this can not be easily done, since no human model has access to the ground correctness of any recommendation. Otherwise, at time step $T$ any of the following should occur:  (a) There is  indifference between accepting the recommendation generated at $T$ and any of the recommendations generated at steps 1...$T-1$, or (b) The expected loss at step $T$ is smaller than the  loss of accepting any of the previous  recommendations generated at time steps $1...(T-1)$.

Thus, for the process to terminate at step $T$, it must hold that
\begin{align*}
    &[h_{acc}(s_T)\prod^{T-1}_{j=1}(1-h_{acc}(s_j))L(s_T) ]\leq\\&[\sum^{T-1}_{i=1}h_{acc}(s_i)L(s_i)\prod_{j-1}^{i-1}(1-h_{acc}(s_j)]
\end{align*}
Thus, to bound the loss near to zero, for any arbitrarily small real number $\epsilon>0$, it should hold that:
\begin{align*}
   &[h_{acc}(s_T)\prod^{T-1}_{j=1}(1-h_{acc}(s_j))L(s_T)] \leq \epsilon \leq \\&[\sum^{T-1}_{i=1}h_{acc}(s_i)L(s_i)\prod_{j-1}^{i-1}(1-h_{acc}(s_j)] 
\end{align*}
\noindent Taking the first inequality, the objective is to reach a state $s_T$ where the following \textit{termination condition}  holds: 
\begin{align}
     L(s_T) \leq \frac{\epsilon}{h_{acc}(s_T)\prod^{T-1}_{j=1}(1-h_{acc}(s_j))}
\end{align}
The denominator of the right inequality term involves the multiplication of numbers less than 1. Thus, this ratio at a time step $T$ shall be larger than $L(s_T)$. 

The following theorem shows that with the objective to minimize $\mathcal{L}_T$, the termination condition specified by the inequality (1) guarantees the termination of the iterative process. 

\begin{theorem}
    The iterative decision-making process with the stated termination condition terminates for any $\epsilon \geq 0$.
\end{theorem}

The proof of this theorem (provided in Appendix B) shows that the upper bound of the difference in loss in two subsequent loss function updates from $t-1$ to $t$ and from $t$ to $t+1$  increases by a factor that is  greater or equal to 1, and is inversely proportional to the probability of the human to reject the recommendation made in state $s_t$. %\footone{The full submission with the appendices can be accessed here: \url{https://drive.google.com/file/d/1lYKqDOZks6CcVOQfgtYj_POmlqOus9jy/view?usp=sharing}
%It can be proved that under a more general condition, where $\frac{h_{acc}(s_{t+1})}{h_{acc}(s_{t})} \geq k \in \mathbb{R}^+$, the process terminates when $\epsilon \geq k /(1-k)$.
%%%%%%%%%%%%%%%%%%%%%%%%%%%%%%%%%%%%%%%%%%%%%%%%%%%%%%%%%%%%%%%%%%%%%%%%

The additional importance of this theorem is that, any choice of $\epsilon$ will terminate the process, independently of whether the probability of user acceptance to recommendations increases. Choosing the value of $\epsilon$ we can tune the effectiveness of the process in terms of the maximum rounds of iteration. 
However, it must be noted  that termination does not imply $successful$ termination: The process may terminate either by accepting an erroneous recommendation that satisfies human constraints (recall that the human has no information on the correctness of any AI recommendation) or by not accepting a recommendation that is both correct and satisfies the constraints.

\section{Human-centric reflective architecture (HCRA)}

\begin{figure*}[t]
\centering
\includegraphics[width=0.75\textwidth]{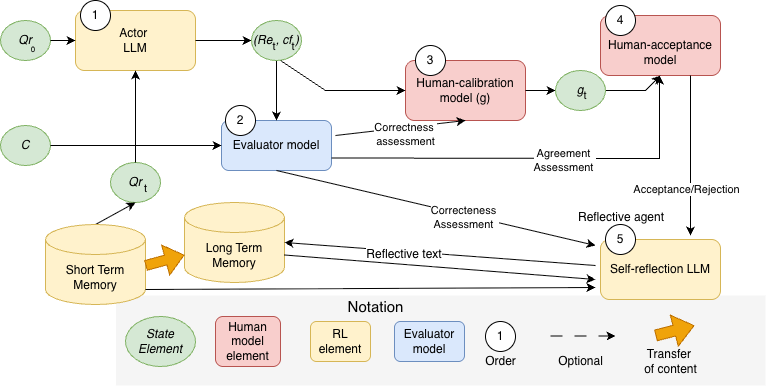} 
\caption{The overall human-centric reflective architecture. }
\label{fig:arch}
\end{figure*}

As it is shown in Figure \ref{fig:arch}, the proposed human-centric reflective architecture consists of five  functional components: (a) those of  the \textit{reflective agent} (shown in yellow), i.e., the actor and the self reflection LLM, 
(b) those that constitute the \textit{human behavior model} (shown in red), i.e. the  human calibration model and the human acceptance model, and (c) the \textit{evaluator} (indicated in blue). The long-term and short-term memories are experience buffers for the agent,  with different time horizons. 

Subsequently, we describe each component and explain their dependencies and their  joint operation. Numbers  in Figure \ref{fig:arch} specify the order of components' execution in each iteration of the decision-making process, based on their input requirements and dependencies, as specified by arrows. %Details on components' input features are described in detail in Appendix A.1.2.

\textbf{Long term memory}: The long term memory stores a long history of past interactions. Specifically, the long term memory is updated with the content of the short term memory, when the iterative process per request terminates. This creates a comprehensive knowledge base that informs future decision-making.

\textbf{Short term memory}: The short term memory stores the  interactions regarding the last request. Specifically, the short term memory stores per iteration the original question, actor recommendation, evaluator assessments, the actor confidence, and generated reflective text.

When generating  reflective text or recommendations, the self-reflection component and the actor, respectively, retrieve stored  interactions to identify specific patterns enabling  targeted feedback generation and recommendations. As already pointed out in Section 2,  memory components facilitate addressing specific deficiencies in making recommendations.

\textbf{Actor}: This is an LLM that serves as the primary recommendation generator in our reflective decision-making process. At each iteration $t$, the actor receives as input the $Qr_t$, which is the initial $Qr_0$ enhanced with the three most recent interactions fetched from the short term memory. It provides the AI recommendation $Re_t$, along with the AI confidence, $cf_t$. 
The template for the actor prompt is specified in Figure 2(top).

\begin{figure}[t]
\centering
\includegraphics[width=0.8\columnwidth]{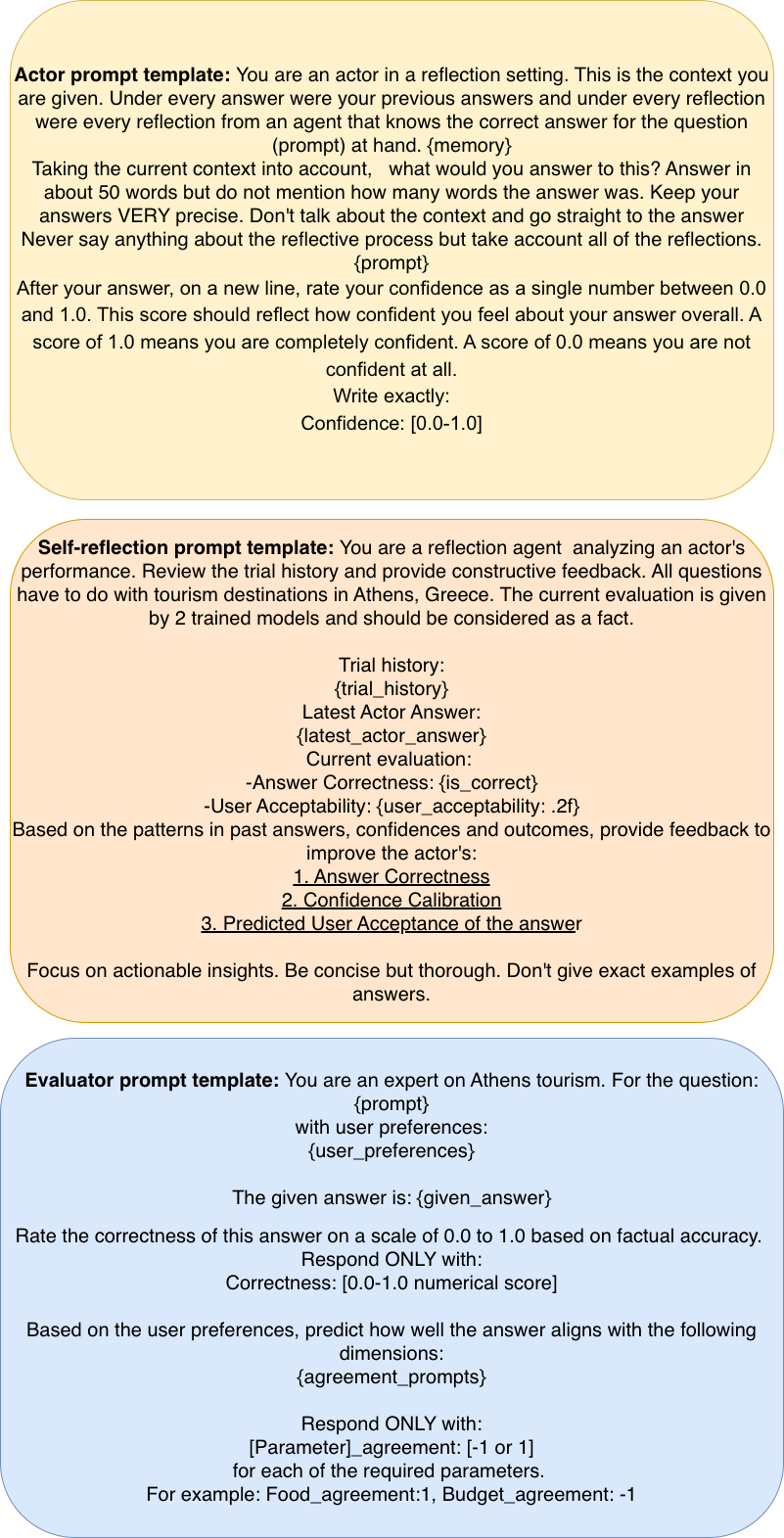} 
\caption{Prompt templates}
\label{fig:prompts}
\end{figure}

\textbf{Evaluator}: The evaluator is an LLM that assesses  the correctness and the agreement  of the recommendation $Re_t$. The template for the evaluator prompt is provided in Figure 2 (bottom). In terms of correctness, the agent evaluates the factual accuracy of the $Re_t$ in relation to $Qr_t$, assigning a score of 1 to $\widehat{{Corr_{t}}}$ if it assesses that the recommendation is correct; otherwise, it assigns -1. To assess agreement, the evaluator focuses exclusively on  specified constraints $Cs$. For each constraint, the evaluator assigns a score of 1 if the $Re_t$ is assessed to align with it, otherwise it assigns -1 . The overall agreement $\widehat{{Aggr_{t}}}$ is determined to be 1 if assessments in relation to individual constraints  are 1; otherwise, it is -1.

%\begin{figure}[t]
%\centering
%\includegraphics[width=0.8\columnwidth]{EPrompt.png} 
%\caption{The Evaluator prompt template}
%\label{fig:arch}
%\end{figure}

\begin{figure}[t]
\centering
\includegraphics[width=1.0\columnwidth]{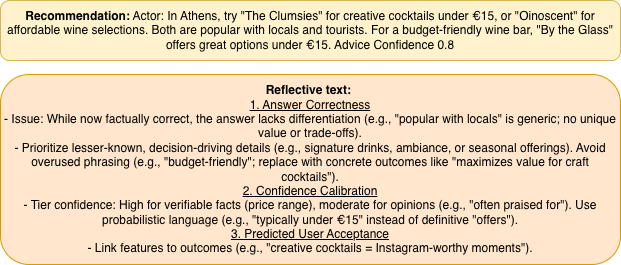} 
\caption{Example of reflective text for an actor-generated recommendation.}
\label{fig:refl_ex}
\end{figure}

\textbf{Self-reflection}: This LLM analyzes the actor’s past recommendations and generates reflective text (linguistic feedback) to guide the generation of the next AI recommendation. It exploits (a) the long term memory, fetching per historical request: the question, the final actor recommendation, and the final reflective text, (b) the interactions stored in short term memory (if any), and (c) the last $Re_t$ provided by the actor, $\widehat{{Corr_{t}}}$ provided by the evaluator, and $h_{acc}(s_t)$ provided by the human acceptance model. The self-reflection model prompt is specified in Figure 2 (middle). The feedback targets the improvement of three key aspects of the actor’s behavior in iteration $t+1$: The assessed correctness of the recommendation $\widehat{{Corr_{t+1}}}$, the actor confidence $cf_{t+1}$, and the probability $h_{acc}(s_{t+1})$ of the human  to accept the recommendation.
An example of reflective text along with the corresponding recommendation is provided in Figure \ref{fig:refl_ex}.

%\begin{figure}[t]
%\centering
%\includegraphics[width=0.8\columnwidth]{SRPrompt.png} 
%\caption{The self-reflection prompt template}
%\label{fig:arch}
%\end{figure}

\textbf{Human acceptance model}: This is part of the  
human behavior model, and realizes the function $h_{acc}$. This model simulates  humans evaluating $Re_t$ and outputs the probability that humans accept the AI recommendation at time step $t$. It takes as input  the human-calibrated confidence of $Re_t$, $g_t$, the assessment $\widehat{{Aggr_t}}$, together with human demographic features.

\textbf{Human calibration model}: This model modifies the actor’s confidence $cf_t$, and provides the human-calibrated confidence $g_t$ for $Re_t$. The goal of training this model is to identify the optimal confidence range that increases $h_{acc}(s_t)$, with respect to the correctness assessed $\widehat{{Corr_t}}$ of  $Re_t$, reflecting the human expectation from the generated recommendation.

The proposed approach is human-centric as evidenced by the following: First, the agent action (the reflective text) depends on  the assessed probability that  the AI recommendation is acceptable by humans, and historical human-AI interactions. Second, the formulated reward of the agent depends on human-calibrated expectations ($g_t$), needs ($Qr_t$) and specified constraints ($C$), and third, the objective aims at  minimizing the final human loss $\mathcal{L}_T$ over a horizon of $T$ iterations.

\subsection{The overall human-calibrated decision-making process} 
As specified in Figure \ref{fig:arch} by the ordering of components execution, initially, the human provides the question $Qr_0$, with constraints $Cs$ that the final decision must satisfy. While $Cs$ remain constant throughout the iterative process, in subsequent iterations $t>0$, $Qr_t$  is  enhanced with the three most recent interactions stored in short term memory, and the reflective text.  

Given $Qr_t$ and $Cs$, an iteration of the reflective process involves the execution of the following components in order:

(1) The actor LLM generates the recommendation $(Re_{t}, cf_{t})$, given $Qr_t$. 

(2) The evaluator takes this recommendation, $Qr_t$ and $Cs$, and provides $\widehat{{Corr_{t}}}$ and $\widehat{{Aggr_{t}}}$. While $\widehat{{Corr_{t}}}$ is exploited by the  human-calibration model and the self-reflection model, the $\widehat{{Aggr_{t}}}$ is exploited by the human acceptance model. 

(3) The human calibration model provides the human-calibrated confidence $g_t$, which is exploited by the human acceptance model. 

(4) The human acceptance model predicts the recommendation acceptance probability $h_{acc}$, which is provided to the self reflection model. 

(5) The self reflection model, given $Re_t, \widehat{{Corr_{t}}}, h_{acc}(s_t)$, the interactions stored in the short term memory and the last interactions per request stored in the long term memory, generates the reflective text.  

(6) The short term memory is then updated to include the new interaction, comprising $Qr_t, (Re_t, cf_t), \widehat{{Corr_{t}}}, \widehat{{Aggr_{t}}}$, and the generated reflective text.

In this work, humans are represented by behavioral models. While humans can theoretically actively participate in the loop by providing feedback on the actor's recommendations—thereby tuning \(g_{t}\) and the acceptance probability \(h_{acc}(s_t)\)—this study utilizes a generalized human behavior model trained on real-world human data, as detailed in the next section. Actively involving human participants and leveraging live interventions during the iterative process are beyond the scope of this work and are reserved for future research.

%%%%%%%%%%%%%%%%%%%%%%%%%%%%%%%%%%%%%%%%%%%%%%%%%%%%%%%%%%%%%%%%%%%%%%%%
\section{Training the models
}

The training process involves the training of the human behavior models, and the training of the reflective agent. 
The training of the human behavior models follows a two-stages training process: We first train the human acceptance model, and then we exploit this well trained model to train the human calibration model.

\begin{figure*}[t]
\centering
\includegraphics[width=0.7\columnwidth]{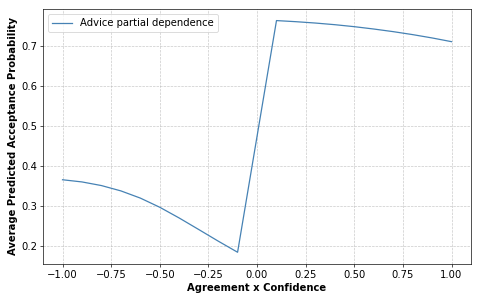} \includegraphics[width=0.68\columnwidth]{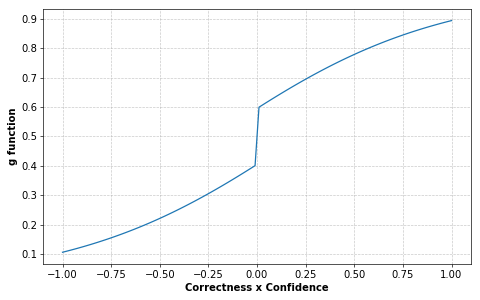} 
\caption{The effect of (left) agreement on $h_{acc}$  and (right) of correctness on human-calibrated confidence  $g(s_t)$.}
\label{fig:acc}
\end{figure*}

\textbf{Human acceptance model}: This model is a three-layer fully connected neural network (10×24, 24×12, and 12×1) with ReLU activation. 
It is trained using binary cross-entropy loss, along with the Adam optimizer \cite{kingma2017adammethodstochasticoptimization} and early stopping to prevent overfitting. To train the model we construct a dataset using data from the dataset described in \cite{10.1145/3514094.3534150}. The constructed dataset is inclusive with  interactions from four  tasks (art, cities, sarcasm, census) and many humans from multiple countries and continents. Important details about the dataset, the features used, and its use for the training of models are included in Appendix A. 

Towards a general and inclusive human behavior model for the purposes of this study, we use the entire constructed dataset. The human acceptance model achieves a ROC-AUC score of 0.78 on the test set of unseen interactions. Although differing training configurations prevent a direct comparison, this result closely aligns with the 0.81 ROC-AUC baseline reported for the activation model in \cite{10.5555/3600270.3600559}.

Figure \ref{fig:acc} (left) presents a partial dependence plot for the human acceptance model. The plot shows the average predicted acceptance probability on the test set of unseen interactions when agreement is fixed across samples. Positive (negative)  values in x-axis indicate agreement (resp. disagreement) with constraints. The absolute values in the x-axis indicate the ``raw" confidence ($cf_t$) of the AI recommendation in time step $t$. The plot reveals that humans  accept (reject) recommendations that respect (resp. disagree with)  their constraints.  The behavior of the model is intuitive, considering that  humans evaluate recommendations mainly based on agreement with constraints, possessing no information regarding correctness. Details about the dataset are in Appendix A.1.

\noindent\textbf{Human calibration model:} %\GV{VERY poor description: Does not explain WHY the formulations are as they are - in contrast to the originals, and does not also justify the choices - It merely describes WHAT}:\\
Similarly to how AI confidence is treated in \cite{10.5555/3600270.3600559}, we consider the inverse sigmoid of
$h_t=(\widehat{Corr_t} \cdot cf_t)$ given the recommendation $(Re_t, cf_t)$ in iteration $t$ and the assessed correctness by the evaluator. Thus, the human-calibration confidence model optimizes the function $g: [0, 1] \times \{-1,1\} \rightarrow [0, 1]$ 
\noindent that maps ($cf_t$,  $\widehat{Corr}_t$) to [0,1]. 
The function $g$ is of the following form:
\[
   g_t= g(cf_t, \widehat{Corr}_t) 
 \]
 \[=  \frac{1}{1 + e^{-sign(\widehat{Corr}_t)(\alpha \cdot sign(\widehat{Corr}_t )\cdot h_t + \beta)}}\]
 \[= \frac{1}{1 + e^{-\widehat{Corr}_t(\alpha \cdot cf_t + \beta)}}
\]

\noindent where $\alpha, \beta \in \mathbb{R}_{\geq 0}$. In the case where $(\alpha, \beta)=(1,0)$  then $g$ is the sigmoid function on $h_t$. Similarly to \cite{10.5555/3600270.3600559}, while $\alpha$ modulates the rate at which  $g$ increases as the confidence of the recommendation increases, $\beta$  adjusts the minimum $g$. Providing the human calibration model with the correctness assessment $\widehat{Corr}$ is intuitive, since humans also consider correctness when assessing the confidence and determining their subsequent reaction to the AI proposal.

Given the well-trained human acceptance model and the dataset used for training that model, we optimize the parameters $(\alpha, \beta)$ of $g$ to minimize the expected loss:
\[
(\alpha^*, \beta^*) = \arg\min_{\alpha, \beta} \; \mathbb{E}_t[-\log\big(h_{\text{acc}}(s_t)\big)]
\]

We finally use $|\alpha|$, $|\beta|$. Using this loss function we aim to penalize the model more when $h_{acc}(s)$ is low, rather when $h_{acc}(s)$ is high. In this way, we encourage the model to adjust its confidence in order to increase the likelihood of the user to accept the recommendation given the human-calibrated confidence. Given the optimized $g$, the percentage of accepted recommendations to the total number of interactions in the test dataset is 51\%, compared to 49\% when the confidence is not human-calibrated.

Figure \ref{fig:acc}(right) illustrates the behavior of the trained human calibration model, mapping the relationship between the assessed correctness of recommendations and AI confidence values. Specifically, the model scales up the confidence scores for correct recommendations (represented by positive x-axis values) within moderate AI confidence intervals. This indicates that the model successfully identifies a specific confidence range where users are most receptive to accurate suggestions. Conversely, incorrect recommendations (negative x-axis values) consistently yield low scores \((<0.4)\), demonstrating that the calibration model effectively penalizes highly confident but incorrect AI outputs.

\textbf{Self-Reflection model:}
HCRA operates according to the refle-xion-based paradigm that enables language agents to provide linguistic feedback supporting the generation of correct advices.  Although the actor serves as the primary recommendation generator, the  self-reflection model guides the process providing reflective text, towards  enhancing the efficiency of the actor and minimizing the loss of the human.  % with configurable temperature settings to produce tourism advice along with confidence scores extracted from token-level log probabilities. 
%, evaluating both factual correctness against ground truth and alignment with user preferences across multiple criteria (food, transportation, accommodation, activities, shopping, nightlife, and budget preferences).
%The Human Calibration Model (g function) transforms the Actor's raw confidence scores based on correctness feedback using the previously defined logistic transformation. The Human Acceptance Model predicts the probability that a human user would accept the Actor's advice, taking as input the calibrated confidence from the g function along with demographic features and preference alignment measures to output acceptance probabilities that drive the learning process. 
The  self-reflection LLM model generates linguistic feedback, crucially exploiting the short and long term memories which accumulate experience of past interactions. This  experience enables the AI agent to be increasingly effective to recommendation and feedback provision  considering reward signals and patterns across similar scenarios. %, whereas the actor, leveraging the feedback found in the short-term memory, can make informed policy steps. \GV{Specify what is added in the long-term experience in any iteration - done}

%The reward signals   provided to the agents and the human, as well as the objective of the learning process and the termination condition are as specified in the problem formulation section.  

%where $\varepsilon = 0.01$ is a constant, $f_{\text{acceptance}}(g_I)$ is the current acceptance probability, and $\prod_{j=1}^{I-1}(1 - f_{\text{acceptance}}(g_j))$ represents the cumulative rejection probability from previous iterations. 

%%%%%%%%%%%%%%%%%%%%%%%%%%%%%%%%%%%%%%%%%%%%%%%%%%%%%%%%%%%%%%%%%%%%%%%%
\section{Experiments}

We evaluate our framework within the domain of touristic decision-making. This environment can be highly critical due to strict human requirements and constraints. These include budget limits, accessibility needs, dietary restrictions, and rigid transportation schedules tied to specific events. Furthermore, safety constraints are paramount, as recommendations must actively prevent users from navigating unsafe areas.

Experiments aim to show that: (a)  HCRA-driven decision-making is more effective than a reflective agentic architecture without models of human behavior; (b) the human-centric aspects of HCRA play a significant role in effectiveness of the collaborative approach; (c) the proposed architecture facilitates high-quality recommendations when the actor is able to take advantage of an exploratory response-generation process, without being overly non-deterministic; (d) the role of long-term memory is crucial for the agent to be effective in answering new requests.  

We measure the quality of recommendations in terms of successful terminations, depending on factual correctness (which neither the AI nor the human models can access) and agreement to constraints. A termination is considered successful if at the final time step the acceptance probability is greater than $0.5$,  the recommendation is correct and in agreement to the constraints. 

Effectiveness is measured by means of (a) the number of iterations needed for the iterative process to terminate, (b) the percentage of successful terminations to the total number of terminations. 

Furthermore, we provide human loss at the final iteration, demonstrating that behavioral models successfully minimize human loss while simultaneously enhancing recommendation effectiveness.   

We have formed 32 questions at different levels of difficulty, depending on the scarcity of information required to form a decision. The iterative process for answering an individual question is a ``trial". Each trial requires a number of iterations until reaching the terminating condition, and can terminate either successfully or not.  We execute 10 independent runs, each one involving one trial per question (i.e., 32 trials per run). At the beginning of a run, we randomly sample human demographic features (age, gender, socioeconomic status, education, programming experience, and AI preference scores)  from the human-behaviour models training dataset. These features apply to all trials performed in the run. In addition, human constraints are randomly selected per trial by multiple criteria for food, transportation, accommodation, activities, shopping, nightlife, and budget attributes. 
In addition to these 32 questions we have formed 10 questions that are either similar to some of the 32, or more complex, in the sense that their recommendation involves the combination of multiple decisions. Using these 10 questions, we evaluate the effectiveness of HCRA to answer questions exploiting  historical interactions stored in the long-term memory. In our case these historical interactions are gathered after responding to the 32 questions.  All questions are specified in Appendix C. %For each episode, the Actor generates an initial recommendation, which is then evaluated by the Evaluator across correctness and preference alignment dimensions to produce individual agreement scores for each relevant criterion. The Actor's confidence is transformed through the g function based on correctness feedback, and the resulting calibrated confidence is fed into the Human Acceptance Model along with the randomly sampled demographic features and preference alignment measures to predict user acceptance probability. The Reflexion agent then analyzes the complete trial history, including Actor recommendations, confidence scores, evaluation outcomes, and acceptance predictions, to generate constructive linguistic feedback for improving subsequent recommendations. This iterative process continues within each episode, with trial memory accumulating context while the framework maintains long-term memory across episodes within each trial to preserve successful and unsuccessful patterns.

The implemented HCRA uses the DeepSeek-V3-0324 model playing the roles of the actor, evaluator and self-reflection model. 
We vary the temperatures of the actor and of the evaluator, as 
specified in different experimental settings, to evaluate the exploratory nature of the actor and the effect of noisy evaluator assessments.
The $\epsilon$ parameter of the iterative process termination condition has been set to 0.01. 

\subsection{Experimental results}

First, to evaluate the impact of the actor model temperature parameter and investigate the role of the actor temperature to steer the quality of decision of HCRA, we performed experiments using different actor temperature values. We report on the effectiveness of HCRA in terms of success rate and average number of iterations (Avg Iters), and we provide the average  loss in the last iteration (Avg Loss) of the iterative process.

\begin{table}[]
\centering
\caption{Performance with different actor temperatures}
\label{tab:temp_analysis}
\resizebox{0.475\textwidth}{!}{%
\begin{tabular}{cccc} \toprule
Temperature & Success Rate (\%) & Avg Iters & Avg Loss \\ \midrule
0.1 & 61.6 & 4.9 $\pm$ 1.64 & 0.373 $\pm$ 0.276\\
0.5 & 55.0 & 5.02 $\pm$ 1.69 & 0.353 $\pm$ 0.178\\
1 & 58.4 & 4.78 $\pm$ 1.81 & 0.334 $\pm$ 0.135 \\
1.5 & 57.8 & 4.89 $\pm$ 1.80 & 0.324 $\pm$ 0.160\\
1.9 & 29.1 & 4.98 $\pm$ 1.8 & 0.360 $\pm$ 0.334\\
\bottomrule
\end{tabular}%
}
\end{table}

Table 1 summarizes the performance metrics across different configurations, with comprehensive data compiled in Appendix C.3. The empirical results demonstrate that HCRA exhibits notable robustness to changes in actor temperature, maintaining a high success rate across most settings. The architecture struggles only at an extreme temperature of 1.9, where it fails to guide the generation process toward successful outcomes. Notably, setting the temperature below 1.0 increases both the final average loss and the mean number of iterations compared to a baseline temperature of 1.0. Conversely, raising the temperature to 1.5 further minimizes the final iteration loss, though this occurs at the expense of an increased iteration count and a slightly lower overall success rate. 
With a high temperature (equal to 1.9) HCRA is unable to steer the process towards successful responses, while both, the average loss and the average number of iterations increase. These results show that temperatures higher or lower than 1 fail to steer decision making effectively. For temperatures lower than 1, the actor becomes more rigid and struggles to incorporate the received feedback, resulting in an increased number of iterations and high average loss. Similarly, for temperatures higher than 1, the actor struggles to respond successfully due to non-determinism, even after receiving the reflective text, resulting in an increased number of iterations. %and high average loss. 
Therefore temperature equal to 1 allows the agent to steer the decision making process effectively, taking  advantage of an exploratory process for the generation of responses. These conjectures are further supported by results reported in Figures \ref{fig:id7-1}, \ref{fig:id3-1}, \ref{fig:id2-1}, \ref{fig:id1-1}, \ref{fig:id8-1}, \ref{fig:id5-1}, \ref{fig:id4-1} in Appendix C.3. 
%As shown in Figure \ref{fig:id2-1} and discussed in the next paragraph,  the highest percentage of successful responses is observed in iterations 3 and 4, where the actor temperature is set to 1.  Based on this performance, the actor temperature is set equal to 1 for all subsequent experiments.

\begin{figure}[]
\centering
\includegraphics[width=1.0\columnwidth]{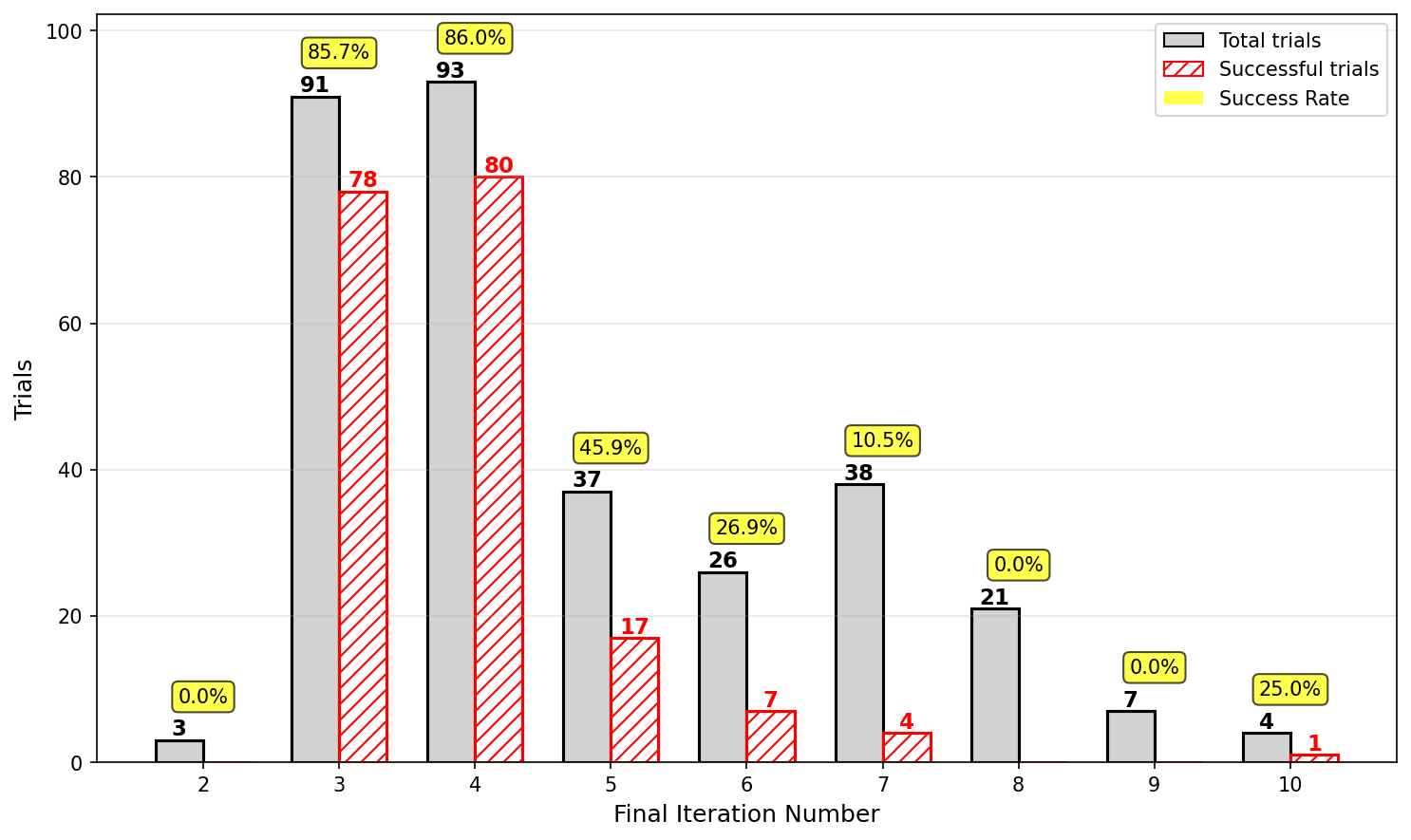} 
\caption{Distribution of total vs successful trials (actor temperature=1.0).}
\label{fig:term_distr}
\end{figure}

To delve into the results for actor temperature equal to 1, we provide results regarding the effectiveness of the process in terms of the successful trials and the number of iterations that they require: Figure~\ref{fig:term_distr} provides the proportion of successful trials  (red bars) to the total number of trials (gray bars) in relation to the  required number of iterations (x-axis). Results show the effectiveness of HCRA: It achieves an overall success rate of 58.4\%, with the majority of trials (57.5\%) reaching termination at the third or fourth iteration. It must be noted that the x-axis starts from 2, given that no trials ended earlier. Notably, 85.7\% of 3-iteration trials and 86.0\% of 4-iteration trials resulted in successful terminations.  The distribution shows a preference for early termination when recommendations meet acceptance criteria, with more iterations (5--10) representing difficult cases requiring extensive refinement of recommendations.
Based on this performance, the actor temperature is set equal to 1 for all subsequent experiments.

\begin{figure}[]
\centering
\includegraphics[width=1.0\columnwidth]{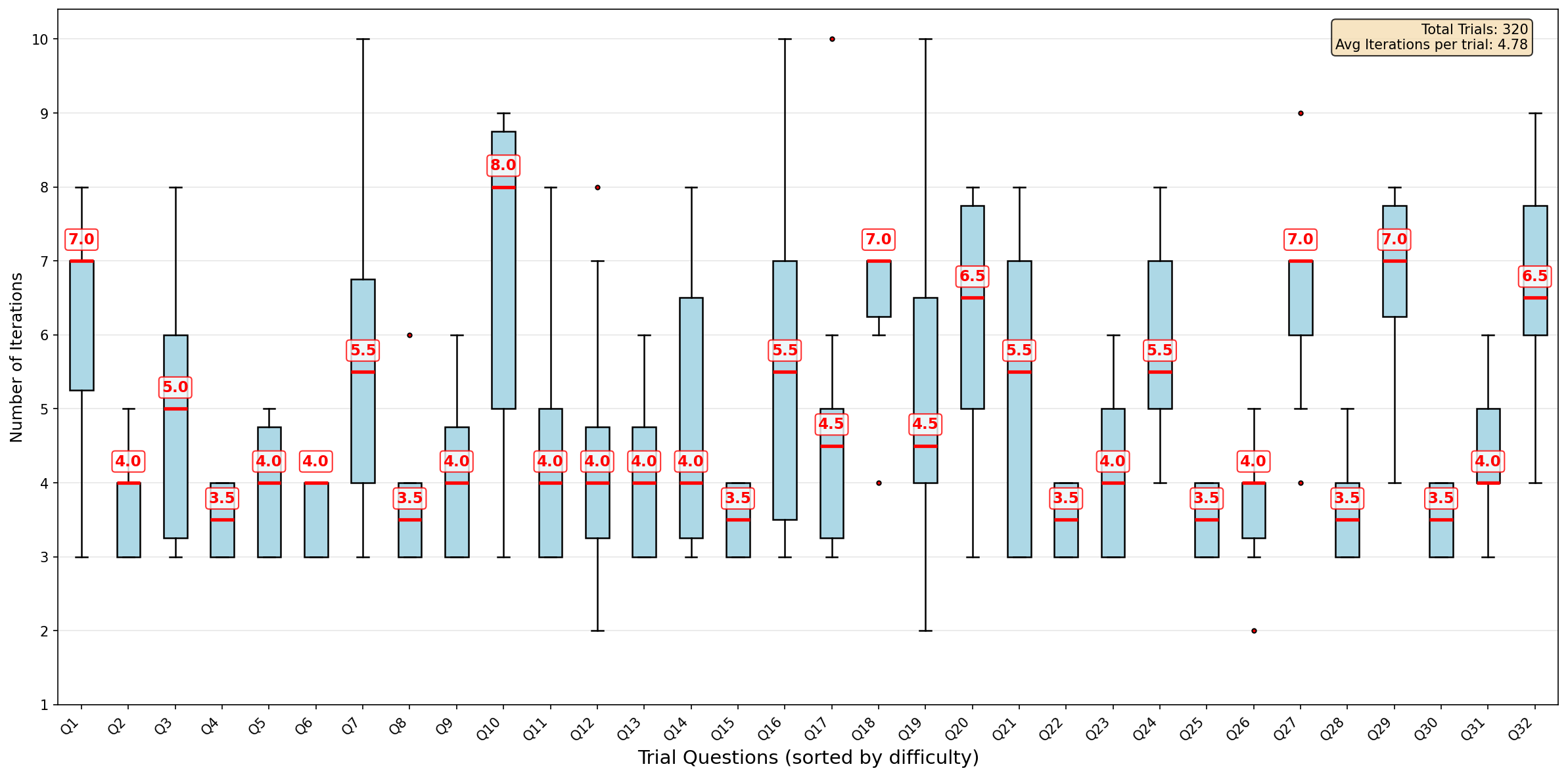} 
\caption{Iterations required per question (actor temperature =1.0).}
\label{fig:av_episodes}
\end{figure}

Focusing on the required number of iterations per question, Figure~\ref{fig:av_episodes} shows the number of iterations required per question (sorted by difficulty left to right) in 10 independent trials and in the form of boxplots. This reveals substantial variation in iteration requirements across different questions, with most of the trials requiring at most 8 iterations and 4 outliers requiring at most 10. The average  number of iterations for all trials is 4.78. The results show consistent behaviour with a median number of 3.5--6.5 iterations for most questions.

\begin{figure}[]
\centering
\includegraphics[width=1.0\columnwidth]{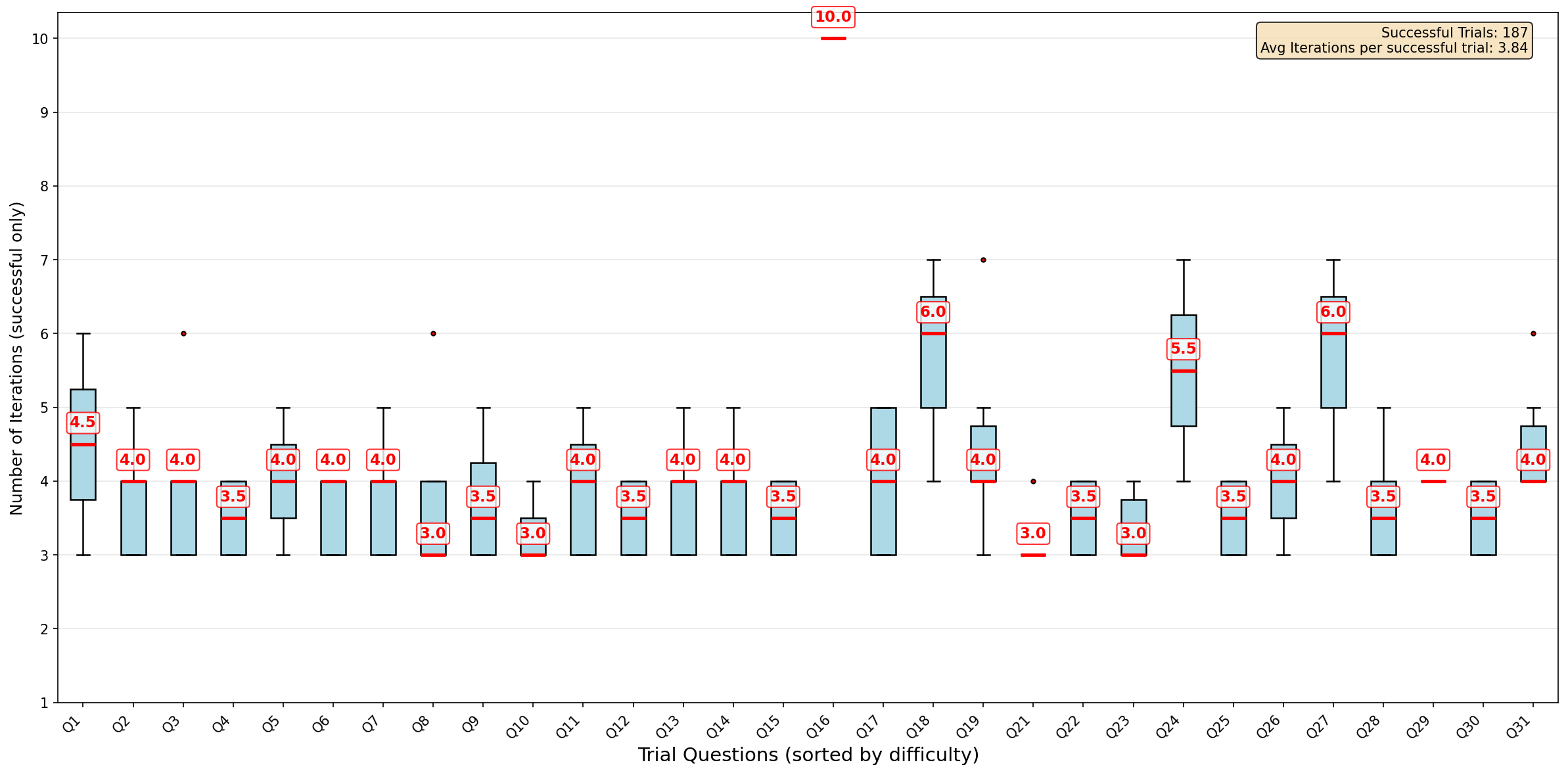} 
\caption{Iterations required for successful termination of questions (actor temperature=1.0).}
\label{fig:succ_term}
\end{figure}

Figure~\ref{fig:succ_term} shows the number of iterations required for the successful trials per question. As shown, these trials require an average number of 3.84 iterations. This is an improvement over the average number of iterations for all trials, suggesting that successful trials tend to terminate  faster than unsuccessful ones, which is quite reasonable considering that unsuccessful trials should imply a kind of difficulty. The median number of iterations across most questions ranges from 3.0 to 4.0, indicating robust HCRA behavior regardless of question difficulty, with only a subset of 4 questions requiring a higher median number of iterations (up to 6.0), and one question reaching the maximum number of 10 iterations.

\begin{table}[]
\centering
\caption{Performance of the trained reflective model. }
\label{tab:learning}
\resizebox{0.475\textwidth}{!}{%
\begin{tabular}{cccc} \toprule
Temperature & Success Rate (\%) & Avg Iters & Avg Loss \\ \midrule
1 & 75 & 4.32 $\pm$ 1.61 & 0.335 $\pm$ 0.254 \\
\bottomrule
\end{tabular}%
}
\end{table}

Table \ref{tab:learning} shows the effectiveness of the reflective process in answering the additional 10 questions, taking advantage of the interactions gathered when it iterated to answer the 32 questions. The results show that HCRA manages to answer these new and complex questions effectively, significantly reducing the average number of iterations (4.32), while significantly increasing the percentage of successful terminations (75\%), compared to the scores achieved when responding without exploiting past experience.

\subsection{Ablation Study}
\label{sec:ablation}
\textbf{The contribution of the human calibration model:} In the absence of the human calibration model, the reflective process relies directly on the AI recommendation confidence $cf_t$. As shown in Table~\ref{tab:g_abl}, this yields degraded performance, compared to the scores reported in Table~\ref{tab:temp_analysis} for actor temperature set to 1.0. As shown, the success rate drops to 53.8\% with an increased average number of iterations (5.02). The average loss value remains relatively stable in this setting compared to those in Table \ref{tab:temp_analysis}. %Detailed results in Appendix C.3 also show that trials in these settings may require a high number of iterations, although with a decreased success rate. %Interestingly, there is not a consistent trend of the overall performance with the increase of the actor temperature: while temperature 1.5 still achieves the highest success rate (70.9\%) and maintains efficient convergence (4.03 iterations in average), temperature 1.85 shows reduced success  rate (64.3\%) despite the improved average number of iterations to successful termination (3.82). , but for the high temperature setting it is slightly higher (0.391) than the corresponding one of HCRA (0.372). 
These results suggest that the human calibration model contributes to the overall HCRA performance, steering the reflective process towards more effective behavior. %, exploiting the exploratory behavior implied in high actor temperature settings to increased performance and reduced human loss.
Detailed results for this setting are shown in Appendix C.3, Figures \ref{fig:id5-1}, \ref{fig:id5-2}, \ref{fig:id5-3}.

\begin{table}[h]
\centering
\caption{Performance without human calibration}
\label{tab:g_abl}
\resizebox{0.475\textwidth}{!}{%
\begin{tabular}{cccc} \toprule
Temperature & Success Rate (\%) & Avg Iters & Avg Iter Loss \\
\midrule
1 & 53.8 & 5.02 $\pm$ 1.73 & 0.334 $\pm$ 0.181 \\
 \bottomrule
\hline
\end{tabular}%
}
\end{table}

\textbf{The impact of an unreliable evaluator:} In relation to the role of the evaluator, Table \ref{tab:noisyeval_abl}  reports on the performance of HCRA with an evaluator that makes unreliable assessments: At each iteration, with probability 0.5, a random noise term drawn uniformly from $[-1,1]$ is added to the evaluator's correctness probability assessment, clipping the result to [0,1]. The result is compared to 0.7 to make the final boolean correctness assessment. The value of the agreement assessment is also changed from 1 to $-1$ or vice versa. The reported results take into account only the original assessments of the evaluator since these reflect the actor's true performance. The significant drop in success rate highlights the crucial role of the evaluator in HCRA: An unreliable evaluator adds noise in the reflective process, failing to steer HCRA to correct recommendations. Detailed results for this setting are shown in Appendix C.3, Figures \ref{fig:id4-1}, \ref{fig:id4-2}, \ref{fig:id4-3}.
\begin{table}[h]
\centering
\caption{Performance with an unreliable evaluator}
\label{tab:noisyeval_abl}
\resizebox{0.475\textwidth}{!}{%
\begin{tabular}{ccc} \toprule
 Success Rate (\%) & Avg Iters & Avg Iter Loss \\
\midrule
15.9 & 4.51 $\pm$ 1.29 & 0.362 $\pm$ 0.269 \\
 \bottomrule
\hline
\end{tabular}%
}
\end{table}

\noindent The HCRA without the human acceptance model is the baseline architecture whose evaluation is reported in the next paragraph. 
\\

\begin{figure}[H]
\centering
\includegraphics[width=1.0\columnwidth]{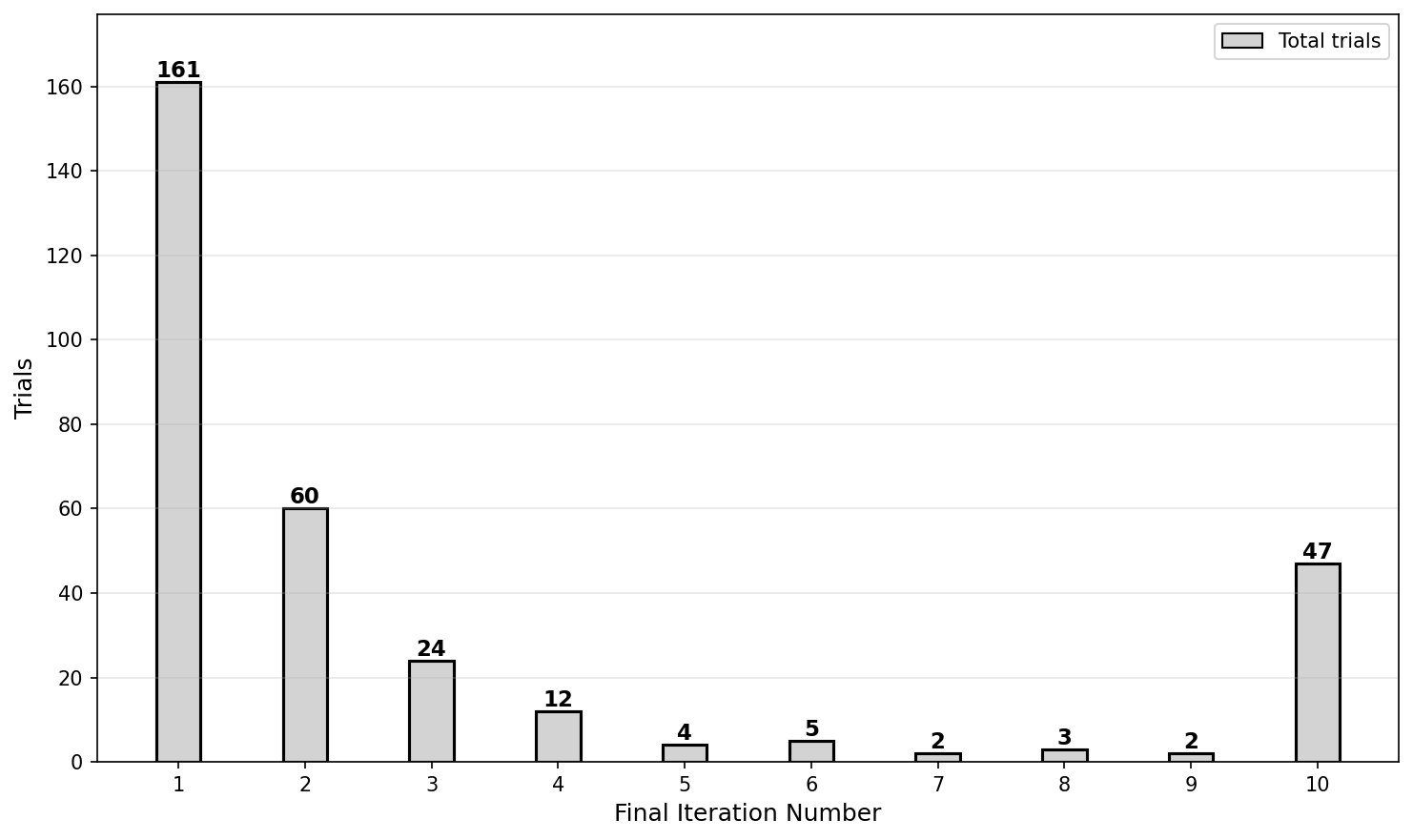} 
\caption{Iterations required for termination when using reflection and factual correctness as termination criterion.}
\label{fig:baseline_iter}
\end{figure}

\textbf{Baseline comparison:} To demonstrate the effectiveness of HCRA, we compare against a baseline system that uses factual correctness as the termination criterion, without exploiting human models and the human loss. Figure~\ref{fig:baseline_iter} reveals the dramatic difference in termination behavior: The baseline system exhibits premature termination with 50\% of trials (161/320) terminating after just one iteration and 69\% completing within two iterations. This contrasts sharply with the results reported for HCRA where the majority (57,5\%) of trials terminate at the third or fourth iteration. The baseline rapid termination is due (a) to the lack of any kind of assessment of recommendations' agreement to human constraints, and (b) to the fact that the probability that the human accepts the recommendation is not exploited in any way towards the final decision. While this approach achieves faster convergence (3.50 iterations on average), it sacrifices human collaboration and recommendation quality with respect to human constraints, while, due to the lack of human models, recommendations are not guaranteed to be acceptable by humans.

%%%%%%%%%%%%%%%%%%%%%%%%%%%%%%%%%%%%%%%%%%%%%%%%%%%%%%%%%%%%%%%%%%%%%%%%

\section{Concluding remarks}
Aiming to improve human-AI collaboration, this work formulates the  human-centric AI-assisted decision-making process as a stochastic game played between the AI agent and  a human. Based on this formulation we  propose HCRA that  leverages models of human behaviour in a reflective process towards refining the recommendations provided by the AI agent according to human objectives and constraints. Experimental results show that the integration of human-calibrated AI models contributes to providing successful recommendations to humans effectively, i.e., in few iterations, favoring explorative behavior and flexibility in responding.  
To ensure utility of the HCRA architecture in specific real-life settings that necessitate the modeling of idiosyncratic human behavior, it needs the following: (a) A highly tuned model of humans' behavior, trained in the domain and context of system use, with data from the humans that  collaborate with the system; (b) a language agent that is able to provide domain-specific recommendations and reflect on them effectively. In settings where fast adaptation to humans is necessary, human behavior models should be trained incrementally to improve their abilities in their context of use, leveraging humans' reactions.

Regarding future work, although we  do not consider long-horizon tasks, we believe that our work can be extended to include such tasks. A recent survey on RL for long horizon interactive LLM agents can be found in \cite{chen2025reinforcementlearninglonghorizoninteractive}. 
Finally, involving  humans in the process, in real-life contexts, is critical to further enhance and assess the utility of the proposed architecture.

%%%%%%%%%%%%%%%%%%%%%%%%%%%%%%%%%%%%%%%%%%%%%%%%%%%%%%%%%%%%%%%%%%%%%%%%

%\begin{acks}
%To Robert, for the bagels and explaining CMYK and color spaces.
%\end{acks}

\bibliographystyle{ACM-Reference-Format}
\bibliography{ref}

\newpage 

\onecolumn

\appendix
\twocolumn
\textbf{APPENDIX }

\section{Dataset and Training of Models}

\subsection{Dataset}
The human behavior models are crucial components of HCRA. In this section we present details for the dataset used for training these models. 

Our starting point is the dataset described in  \cite{10.1145/3514094.3534150}, which includes samples from real human responses. 

For the purposes of our work we selected a subset of samples from that dataset, where the term ``sample" denotes a single interaction comprising a question, an AI recommendation, and a person response. We consider samples  of ``agreement" and ``disagreement", considering ''agreement" as specified in the main part of the article with respect to person constraints. The terms ``acceptance" and ``rejection" indicate the person’s reaction to the provided recommendation. Below we specify how these samples have been selected.

To determine cases of acceptance using this dataset, we computed the absolute difference between the features $response_1$ and $response_2$ (i.e. the difference in confidence between the first and the final person response,  as provided in \cite{10.1145/3514094.3534150}). When this difference is greater than $0.035$ then samples were labeled with ``accept"; otherwise, they were labeled with ``reject". This is justified in \cite{10.1145/3514094.3534150} and further evidence is provided in \cite{10.5555/3600270.3600559} by the fact that this difference indicates whether persons integrate the AI recommendation in their own response. Actually, this happens   when (a) the AI recommendation is provided with high confidence and is opposite to the person’s 
response, and (b) in case the recommendation and the person’s initial response share the same label, the person’s initial response has low confidence and the AI has high
confidence.

To determine cases of (dis)agreement we are using the features $advice$ and $response_1$ as defined in the dataset \cite{10.1145/3514094.3534150}. The sign of each of these features shows whether the corresponding choice is (in)correct in a binary classification task. When both features have the same sign, this indicates that they both, AI and the person, selected the (in)correct answer, which we classify as ``agreement". Different signs indicate that one choice was correct, while the other was incorrect, which we classify as disagreement.

\renewcommand{\arraystretch}{1.3}
\begin{table}[h!]
\centering
\caption{Distribution of samples in the original dataset provided in \cite{10.1145/3514094.3534150} in the different cases of  (non)acceptance and (dis)agreement.}
\resizebox{\columnwidth}{!}{%
\begin{tabular}{cccc}
\toprule
 & Accept & Reject & Total samples \\
\midrule
Agreement & 10,682 & 12,753 & 23,435 \\ 
Disagreement & 8,068 & 3,152 & 11,220\\

\textbf{Total samples} & 18,750 & 15,905 & 34,655\\
\bottomrule
\end{tabular}}

\label{table1}
\end{table}

Table~\ref{table1} presents the number of valid samples in the original dataset in different cases of (non)acceptance and (dis)agreement. A sample is considered valid if it includes  non-null values in the features we use as inputs to the human-behavior model (these features are specified subsequently in A.2). 

\renewcommand{\arraystretch}{1.3}
\begin{table}[h!]
\centering
\caption{Distribution of samples in our balanced dataset.}
\resizebox{\columnwidth}{!}{%
\begin{tabular}{cccc} \toprule

 & Accept & Reject & Total samples \\ \midrule

Agreement & 3,152 & 1,050 & 4,202 \\ 
Disagreement & 1,050 & 3,152 & 4,202 \\

\textbf{Total samples} & 4,202 & 4,202 & 8,404\\
\bottomrule
\end{tabular}}

\label{table2}
\end{table}

As it can be observed, the dataset is unbalanced as far as agreement/ disagreement is concerned, and the proportion of samples in the disagreement/acceptance class is larger to those in the disagreement/rejection. Through a series of experiments, we observed that this strongly influences the predicted human acceptance probability to the provided AI recommendations under the assumptions made in this work. Therefore, we created a new version of the dataset in which, the cases for agreement and disagreement are balanced and there is also a balance in the distribution of samples with labels ``accept" and ``reject" in each case. This has been done in a meticulous manner as it is described subsequently, so as to support in our setting the tendency of persons to evaluate AI responses primarily based on agreement with the constraints, rather than on correctness, given that, as assumed, humans do not have access to any hint or information concerning the ground correctness of AI responses.

To construct such a balanced dataset with the largest possible number of samples, we started from the under-represented case, namely the case disagreement/reject that contains 3152 samples. Using this as a reference, we selected the same number of samples for the agreement/accept case. To allocate  samples to the cases we experimented with different ratios of samples to  the disagreement/reject and agreement/accept cases compared to their complimentary ones, and we adopted a ratio 75/25, where 75\% corresponds to the majority classes (disagreement/reject and agreement/accept) with 3152 samples each, and 25\% to the complementary cases (agreement/reject and disagreement/accept) with 1050 samples each.\\
\begin{figure}[h!]
    \centering
    \includegraphics[width=\linewidth]{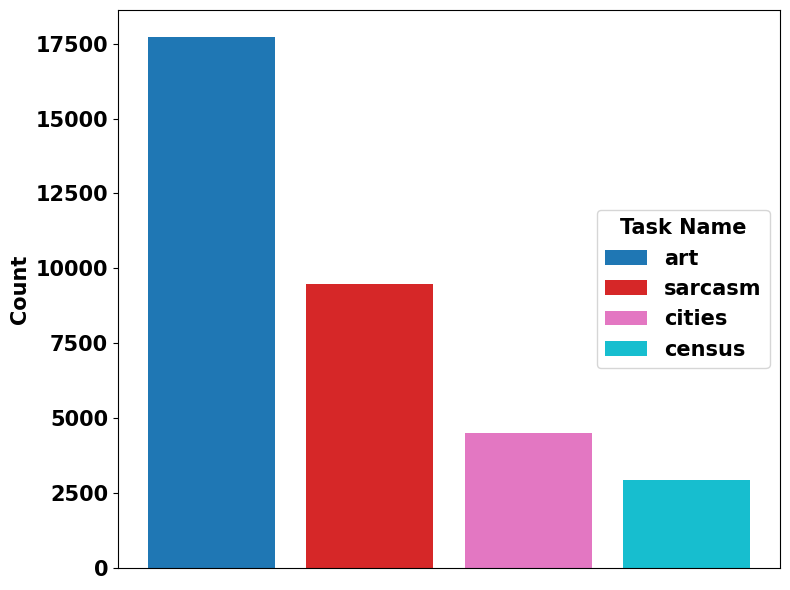}
    \caption{Distribution of samples in tasks, in the original dataset \cite{10.1145/3514094.3534150}}
    \label{fig1}
\end{figure}

\begin{figure}[h!]
    \centering
    \includegraphics[width=\linewidth]{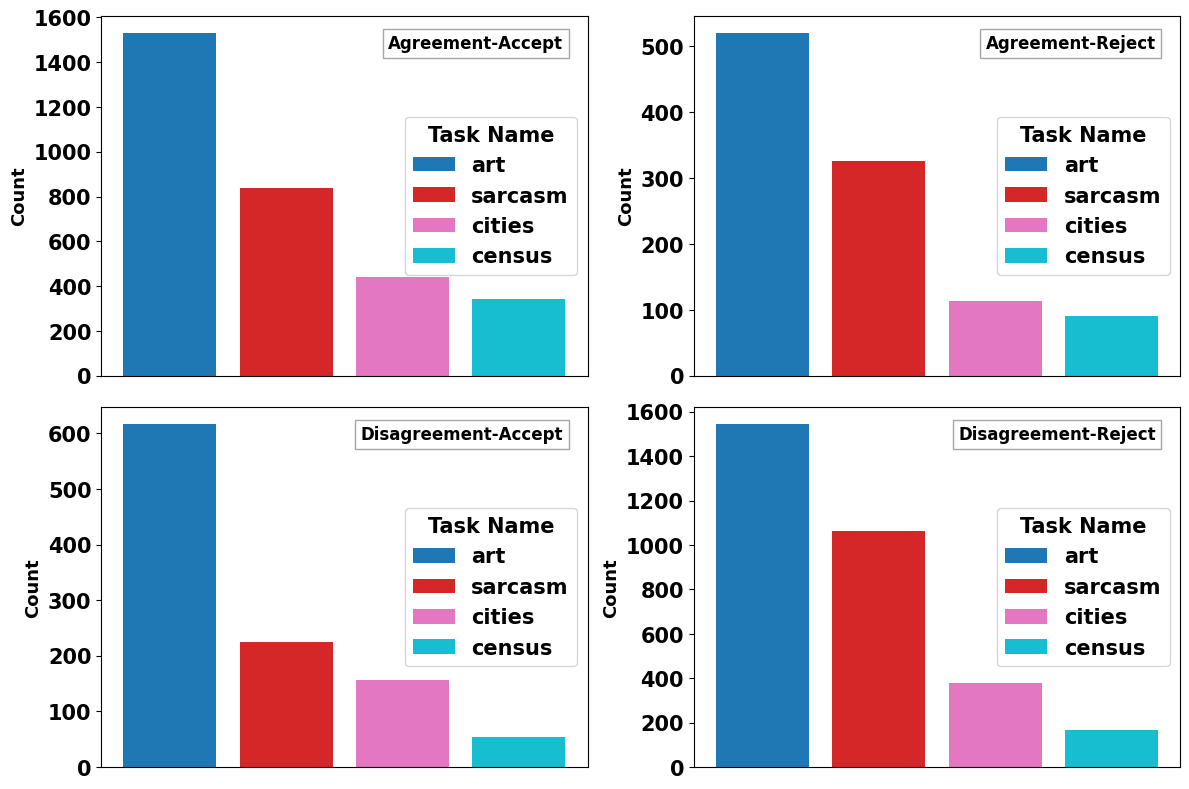}
    \caption{Distribution of samples in tasks, in our dataset }
    \label{fig2}
\end{figure}

\begin{figure}[h!]
    \centering
    \includegraphics[width=\linewidth]{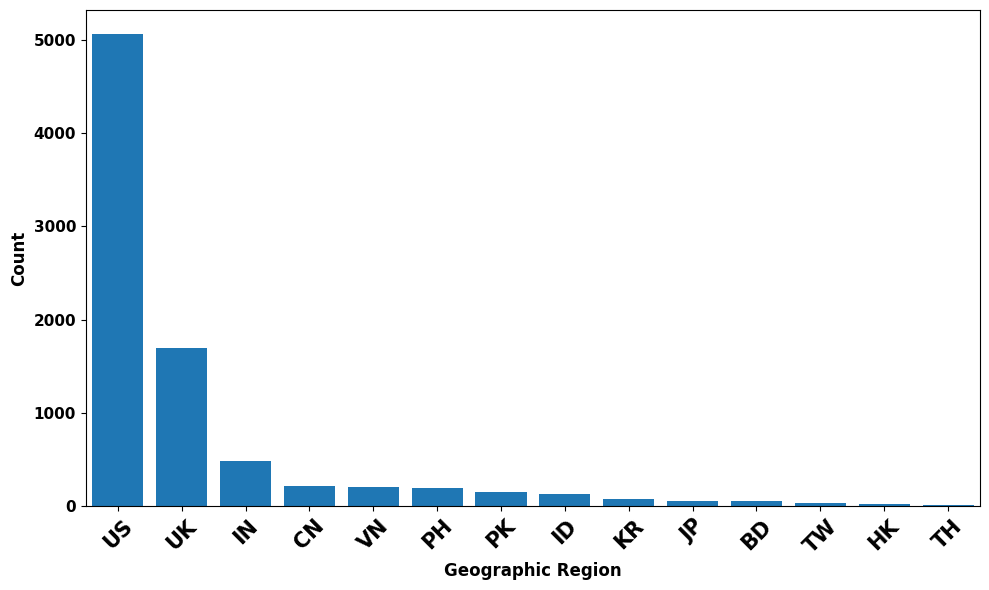}
    \caption{Distribution of samples in geographical regions, in our dataset }
    \label{fig:geo}
\end{figure}

Aiming to aggregate training towards a general and inclusive human-behavior model, we selected samples from all available tasks (art, sarcasm, cities, census). With respect to the 75/25 ratio we ensure that the proportion of samples per task is maintained in all four classes (agreement/accept, agreement/reject, disagreement/accept, and disagreement/reject) as illustrated in Figures~\ref{fig1}, ~\ref{fig2}, and is similar to their proportion in the original dataset. Moreover, we ensured that the final dataset includes samples from persons across multiple countries and continents, thereby preserving diversity in demographic features. The relevant statistics are shown in Figure \ref{fig:geo}.

\begin{table}[h!]
\centering
\resizebox{0.6\columnwidth}{!}{%
\begin{tabular}{c c p{4cm}}
\hline
\textbf{} \textbf{Country} & \textbf{Abbreviation} \\
\hline
United States & US \\
United Kingdom & UK \\
India & IN \\
China & CN \\
Vietnam & VN \\
Philippines & PH \\
Pakistan & PK \\
Indonesia & ID \\
Korea & KR \\
Japan & JP \\
Bangladesh & BD \\
Taiwan & TW \\
Hong Kong & HK \\
Thailand & TH \\
\hline
\end{tabular}}
\caption{Country names and their abbreviations}
\label{table3}
\end{table}

\begin{figure}[h!]
    \centering
    \includegraphics[width=\linewidth]{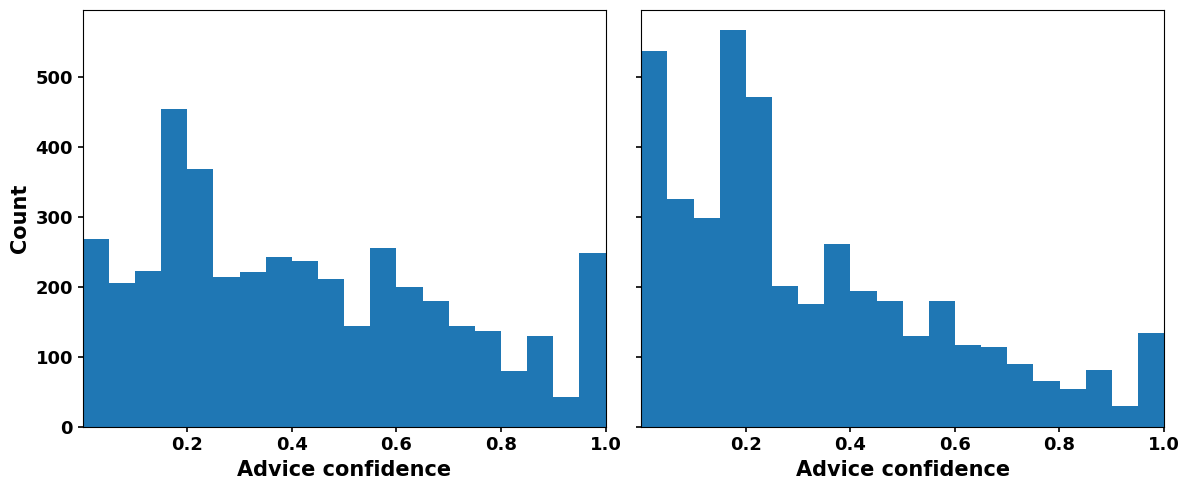}
\caption{Distribution of recommendation confidence in our dataset for the cases where it is accepted (left) and when it is rejected (right)}
    \label{fig4}
\end{figure}

Figure~\ref{fig4} shows the distribution of the AI recommendation confidence for the samples labeled with ``accept" (left) and ``reject" (right). The distributions, although with differences, indicate that all confidence levels are represented in a similar manner in the two classes.

\subsection{
Input features for the human behavior model} 
In this section, we present the features exploited by the human behavior models, i.e., the human acceptance and the human calibration models. 

\renewcommand{\arraystretch}{1.5}
\begin{table}[h!]
\centering
\caption{Input features for human acceptance model}
\resizebox{\columnwidth}{!}{%
\begin{tabular}{ccp{6cm}}
\toprule
\textbf{Feature \#} & \textbf{Value Range} & \textbf{Description} \\
\midrule
1 & $\mathbb{R}_{\geq 18}$ & Age \\
2 & $\{0, 1\}$ & Gender \\
3 & $[1, 10]$ & Self-reported socioeconomic status \\
4 & $[1, 8]$ & Self-reported education level \\
5 & $\{0, 1\}$ & Does the person have prior experience with computer programming? \\
6 & $[-1, 1]$ & Do you think the AI or the average person (without help) can do better on this task? \\
7 & $[0, 1]$ & How often do you use AI systems to aid you in your everyday life and/or at work? \\
8 & $[0, 1]$ & Human calibrated confidence $g_t$ \\
9 & $\{-1, 1\}$ & Agreement assessment $\widehat{Aggr}_t$ \\
10 & $[-1, 1]$ & Feature 8 $\times$ Feature 9 \\
\bottomrule
\end{tabular}}

\label{tab:features_acc}
\end{table}

Table \ref{tab:features_acc} specifies the features used as input to the human acceptance model. Features 1-7 represent person-specific features that remain constant for a person across all interactions, while features 8-10 are state-dependent features that change during the iterative reflective process.  Features 3, 4, 6 and 7 are  as specified in \cite{10.1145/3514094.3534150}, preserving their domains and scales.  More specifically:

\begin{itemize}
    \item \textbf{Feature 3}: Socioeconomic status is a self-assessed feature measured on a 1–10 scale, where 10 indicates individuals who are most advantaged in terms of income, education, and job opportunities, and 1 represents those who are least advantaged.
    \item \textbf{Feature 4}: Education level is a self-assessed feature measured on a scale from 1 to 8. A value of 1 indicates “Don’t know / not applicable,” while higher values correspond to higher education levels, ranging from (2) no formal qualifications, (3) secondary education, (4) high school diploma, (5) technical or community college, (6) undergraduate degree, (7) graduate degree and (8) doctorate degree.
    \item \textbf{Feature 6}: This feature quantifies the person’s confidence in AI performance before the person engages in the task. It takes values in the range $[-1, 1]$, where 1 indicates that AI is expected to perform better and $-1$ that a human would perform better.
    \item \textbf{Feature 7}: This question measures person familiarity with AI. It takes values in the range $[0, 1]$, where 0 indicates never and 1 indicates very frequent use of AI.
\end{itemize}

The input features for the human calibration model are shown in Table \ref{tab:features_hcal}.

\renewcommand{\arraystretch}{1.5}
\begin{table}[h!]
\centering
\caption{Input features for human calibration model}
\resizebox{\columnwidth}{!}{%
\begin{tabular}{ccp{6cm}}

\toprule
\textbf{Feature \#} & \textbf{Value Range} & \textbf{Description} \\
\midrule
1 & $[0, 1]$ & Recommendation confidence $cf_t$ \\
2 & $\{-1, 1\}$ & Correctness assessment $\widehat{Corr}_t$ \\
\bottomrule
\end{tabular}}
\label{tab:features_hcal}
\end{table}

\subsection{Training the human behavior models}

\subsubsection{Human acceptance model}

Exploiting the features specified in our dataset, we applied Min–Max normalization to the ``socioeconomic status" and ``education" features and  z-score normalization for the ``age' feature. The ground truth for the human acceptance model is as described in A.1.1.

It must be noted that during the training of the human acceptance model, neither the human calibration confidence $g_t$ nor the agreement assessment $\widehat{Aggr}_t$ are available. Instead, we use (i) the absolute value of the $advice$ feature from the original dataset in place of $g_t$, and (ii) the derived agreement feature, as defined in A.1, in place of $\widehat{Aggr}_t$. Although the $\widehat{Aggr}_t$ and the agreement feature are not identical, both quantify how closely the AI recommendation is with what the human knows, believes  and/or prefers.

\subsubsection{Human calibration model}

The human behavior model is trained independently and integrated into the overall process when it is well-trained. As a result, the correctness assessment $\widehat{Corr}_t$ required for the training of the human calibration model is not available during training. Therefore, we train the human calibration model 
using the correctness of the recommendation derived directly from the data set. Specifically, if the $advice$ feature in the original dataset has a positive sign, it is considered correct (1), whereas a negative sign indicates incorrect recommendation (-1). 

\onecolumn
\section{Termination of the reflective process } 

\subsubsection{Theorem.} The iterative decision-making process with the stated termination condition terminates.% (a) for any $\epsilon$, when $k \geq 1$ and (b) for $\epsilon \leq \frac{k}{(1-k)}$ when $k < 1 $.
\\

\subsubsection{Proof.}

Let $k = min\{ \frac{h_{acc}(s_{t}+1)}{h_{acc}(s_{t})}, t=0,1,2,\dots \}$.

\begin{align*}
\|L_{t+1}-L_{t}\| \leq  \frac{\epsilon}{h_{acc}(s_{t+1})\prod^{t}_{j=1}(1-h_{acc}(s_j))}-\frac{\epsilon}{h_{acc}(s_t)\prod^{t-1}_{j=1}(1-h_{acc}(s_j))}
\end{align*}

\noindent Given that $\frac{h_{acc}(s_{t+1})}{h_{acc}(s_{t})} \geq k$, then 

\begin{align}
\|L_{t+1}-L_{t}\| \leq \frac{\epsilon - k \epsilon(1-h_{acc}(s_t))  }{k h_{acc}(s_{t})\prod^{t}_{j=1}(1-h_{acc}(s_j))} =
\frac{\epsilon (1-k)}{k h_{acc}(s_t))\prod^{t}_{j=1}(1-h_{acc}(s_j))}+\frac{\epsilon}{\prod^{t}_{j=1}(1-h_{acc}(s_j))} 
\end{align}

(a) When $k \geq 1$ then $(1-k) \leq 0$, so  (2) implies that 
\begin{align*}
\|L_{t+1}-L_{t}\| \leq \frac{\epsilon}{\prod^{t}_{j=1}(1-h_{acc}(s_j))}
\end{align*}
Therefore, $\frac{\|L_{t+1}-L_{t}\|}{\|L_{t}-L_{t-1}\|} \leq \frac{1}{(1-h_{acc}(s_t))}$, i.e. the difference in loss in two subsequent iterations increases  by a factor  greater or equal to 1. Therefore, the process will reach a point $T$ where the terminating condition is satisfied.
\\

(b) When $k < 1$ then $(1-k) < 1$, so   we distinguish two cases:
\\

(b.1) $\frac{\epsilon(1-k)}{k} >1$, if $\epsilon>\frac{k}{(1-k)}$. 

This implies that the first term of the right part of the inequality (1), which is greater than the second term, increases fast as $t$ increases, getting much bigger than 1. 
In this case
\begin{align*}
\frac{\|L_{t+1}-L_{t}\|}{\|L_{t}-L_{t-1}\|}\leq \frac{h_{acc}(s_{t-1})}{h_{acc}(s_t)(1-h_{acc}(s_t))} \leq \frac{1}{k(1-h_{acc}(s_t))}
\end{align*}

(b.2) $\frac{\epsilon(1-k)}{k} \leq 1$, if $\epsilon \leq \frac{k}{(1-k)}$. 

In this case (1) implies

\begin{align*}
\|L_{t+1}-L_{t}\| \leq 
\frac{1}{h_{acc}(s_t))\prod^{t}_{j=1}(1-h_{acc}(s_j))}+\frac{k}{(1-k)\prod^{t}_{j=1}(1-h_{acc}(s_j))}=\\
\frac{1-k+kh_{acc}(s_t)}{(1-k)h_{acc}(s_t))\prod^{t}_{j=1}(1-h_{acc}(s_j))} \leq \frac{1}{(1-k)h_{acc}(s_t))\prod^{t}_{j=1}(1-h_{acc}(s_j))} 
\end{align*}
%It can be proved that under a more general condition, where $\frac{h_{acc}(s_{t+1})}{h_{acc}(s_{t})} \geq k \in \mathbb{R}^+$, the process terminates when $k \geq \epsilon /(1+\epsilon)$. 

Thus, in a similar way to (b.1) it holds that
\begin{align*}
\frac{\|L_{t+1}-L_{t}\|}{\|L_{t}-L_{t-1}\|}\leq \frac{h_{acc}(s_{t-1})}{h_{acc}(s_t)(1-h_{acc}(s_t))} \leq \frac{1}{k(1-h_{acc}(s_t))}
\end{align*}

Therefore, in both sub-cases of (b) the upper bound of the difference in loss in two subsequent loss function updates increases by a factor  greater or equal to 1. Therefore, the process will reach a point $T$ where the terminating condition is satisfied.

\newpage
\onecolumn
\section{Experimental Setting }

\subsection{Questions} 

The following list provides the complete set of questions clustered and ordered by increased difficulty. Clusters of questions of the same difficulty are within subsequent table rules. For each question we specify the available options for the specification of constraints. 
The last cluster of 10 questions are those used for evaluating the learning abilities of HCRA, exploiting gathered experience.

\begin{table}[H]
    \centering
%    \caption{ }
    \scriptsize
    \begin{tabular}{c p{11cm} p{6cm}}
    \toprule
    \textbf{Q.ID} & \textbf{Question} & \textbf{Constraints options} \\
    \midrule
    1 & Can you suggest a family-friendly activity in [location]? & \textbf{Activities}: arts, historical\_sites, indoor\_entertainment, family\_activities, museum; \textbf{Budget}: low, medium, high \\
    
2 & Are there any hostels in [location]? & \textbf{Accom.}: hostels; \textbf{Budget}: low, medium \\
3 & Where can I enjoy traditional [nationality] dishes in this area? & \textbf{Food}: local\_cuisine; \textbf{Budget}: low, medium \\
4 & I want to try some authentic [nationality] street food. & \textbf{Food}: street\_food; \textbf{Budget}: low, medium \\
5 & Are there any good hotels close to [location] Metro Station? & \textbf{Accom.}: hotels; \textbf{Budget}: medium, high \\
6 & What is the most historical place to visit in [location]? & \textbf{Activities}: historical\_sites; \textbf{Budget}: medium, high \\
7 & I’d love to visit a cafe with a magical, fairytale theme. & \textbf{Food}: cafe; \textbf{Budget}: low, medium \\
8 & I’d like to spend some time reading a book while enjoying coffee. Any suggestions? & \textbf{Food}: cafe; \textbf{Activities}: reading; \textbf{Budget}: low, medium \\
9 & Which is the highest-rated vegetarian restaurant in [location]? & \textbf{Food}: vegan; \textbf{Budget}: medium, high \\
10 & Can you suggest a bar where I can enjoy wine or cocktail for under 15 euros? & \textbf{Nightlife}: live\_music, bars, clubs, shows; \textbf{Budget}: low, medium \\
\midrule
11 & Are there any museums near [location] with wheelchair-accessible parking? & \textbf{Transport}: taxi, rental\_car; \textbf{Activities}: museum; \textbf{Budget}: low, medium \\
12 & I want to listen live music while having a drink. Where should I go? & \textbf{Nightlife}: live\_music, shows; \textbf{Budget}: low, medium, high \\
13 & Is there an open-air cinema around [location]? & \textbf{Activities}: cinema; \textbf{Budget}: low, medium \\
14 & Can you suggest a cafe where I can enjoy a view of the [landmark]? & \textbf{Food}: cafe; \textbf{Budget}: low, medium, high \\
15 & Where can I have an affordable breakfast for under 5 euros? & \textbf{Food}: breakfast; \textbf{Budget}: low \\
16 & Which night club in [location] has the most reviews? & \textbf{Nightlife}: clubs; \textbf{Budget}: low, medium \\
17 & Where should I go for a well-known brunch experience in this area? & \textbf{Food}: brunch; \textbf{Budget}: medium, high \\
18 & I need a dog-friendly hotel to stay. & \textbf{Accom.}: hotels; \textbf{Budget}: low, medium, high \\
19 & Where can I watch a football match and have a beer in [location]? & \textbf{Activities}: indoor\_entertainment; \textbf{Nightlife}: pubs; \textbf{Budget}: low, medium \\
20 & I’m looking for a burger restaurant where I can book a table. & \textbf{Food}: restaurants, street\_food; \textbf{Budget}: low, medium \\
\midrule
21 & What’s the nearest station to the [landmark] Museum? & \textbf{Transport}: bus, metro; \textbf{Activities}: historical\_sites; \textbf{Budget}: low \\
22 & What’s the shortest path from [location] to [destination] on foot? & \textbf{Transport}: walking \\
23 & Can I walk to [destination] from [location] in under 15 minutes? & \textbf{Transport}: walking \\
24 & Are there any bowling alleys in [location]? & \textbf{Activities}: indoor\_entertainment, family\_activities; \textbf{Budget}: low, medium \\
25 & What metro lines are available in [location]? & \textbf{Transport}: metro; \textbf{Budget}: low \\
26 & I want to get some souvenirs. Where should I look? & \textbf{Shopping}: locally\_owned\_stores; \textbf{Budget}: low, medium \\
27 & Which cafes in [location] accept NFC payments? & \textbf{Food}: cafe; \textbf{Budget}: low, medium \\
28 & Are there any Italian restaurants with moderately priced services around [city center]? & \textbf{Food}: restaurants, ethnic\_cuisine; \textbf{Budget}: medium \\
29 & I’m looking for a place to eat something and it may be suitable for smokers. & \textbf{Food}: restaurants, street\_food, local\_cuisine, vegan, ethnic\_cuisine; \textbf{Budget}: low, medium, high \\
30 & What is the best way to reach [destination] on foot starting from [location]? & \textbf{Transport}: walking \\
\midrule
31 & Which EV station has the most connectors to charge my electric car? & \textbf{Transport}: rental\_car; \textbf{Budget}: low, medium \\
32 & Which [nationality] restaurant offers a large variety of cocktails? & \textbf{Food}: restaurants, ethnic\_cuisine; \textbf{Budget}: low, medium \\
\midrule
\midrule
1 & Can you suggest a place where I can enjoy a view of historical monuments? & \textbf{Activities}: historical\_sites; \textbf{Budget}: low, medium\\
2 & I’d like to have some time to complete a writing while drinking coffee. Any suggestions? & \textbf{Food}: cafe; \textbf{Budget}: low, medium \\
3 & Are there any well-rated restaurants with moderately priced services in Athens city center? & \textbf{Food}: restaurant, local cuisine; \textbf{Budget}: medium \\
4 & What options for EV stations do exist for charging my electric car in Athens center? & \textbf{Transport}: rental car; \textbf{Budget}: low, medium  \\
5 & Which EV station do you suggest for charging my electric car in Athens center while enjoying coffee in a walking distance cafe with a view in Acropolis? & \textbf{Food}: cafe; \textbf{Transport}: rental car; \textbf{Budget}: low, medium\\
6 & I’m looking for a pizza or Italian restaurant with moderately priced services around Athens city center that it is suitable for smokers.& \textbf{Food}: restaurants, ethnic cuisine; \textbf{Budget}: medium \\
7 & The most known historical places in Monastiraki are usually crowded. What you would suggest for a more relaxed visit that would upgrade my visit of historical places? & \textbf{Activities}: historical sites, museum; \textbf{Transport}: walking; \textbf{Budget}: low, medium \\
8 & Plan for me a walking tour on three historical places in Athens center, starting from a cafe place and with a break for lunch. The total travel distance should not exceed 2 km and should be no boring. Expenses should not exceed 60 Euros in total.& \textbf{Activities}: historical sites, \textbf{Food}: cafe and restaurant; \textbf{Transport}: walking; \textbf{Budget}: low, medium \\
9 & Plan for me a walking tour on two classical museums in Athens center, starting from a cafe place and with a break for lunch. The total travel distance should be weel chair accessible, not exceed 2 km and should be no boring. Expenses should not exceed 60 Euros in total. & \textbf{Activities}: museum; \textbf{Food}: cafe and restaurant; \textbf{Transport}: walking; \textbf{Budget}: low, medium  \\
10 & Which restaurant offers a large variety of cocktails and accepts NFC payments? & \textbf{Food}: restaurant; \textbf{Budget}: low, medium\\
\bottomrule
\end{tabular}
\end{table}

\twocolumn

\newpage

\subsection{Configuration of components}

This section contains details about the hyperparameters and configurations of all components in HCRA.

\subsubsection{Termination hyperparameter.}
Results reported  are from HCRA with the parameter $\epsilon$ set to $0.01$, affecting the termination of the iterative reflective process. 

Higher values of $\epsilon$ are appropriate for decision-making tasks where different issues have to be reconciled and potential complexities in constraints and recommendations should be resolved, thus for tasks requiring a large number of iterations. Small values of $\epsilon $ are appropriate for decision-making in question-answering settings requiring few iterations.
Based on our experimental results, $\epsilon=0.01$ allows for sufficient response refinement, achieving high rates of successful decisions, i.e., of responses meeting all three criteria: acceptance probability exceeding 0.5, assessed correctness, and assessed agreement  with respect to person's needs (request) and constraints.

We also limit the number of iterations up to a maximum number providing a safety net for edge cases requiring an excessive number of iterations. This is not a hyperparameter of the  method, but we specify this here for reasons of comprehensiveness. We have set a maximum of 10 iterations which has been reached only in very few cases, and exclusively during ablation study conditions, indicating that the proposed HCRA rarely requires this high number of iterations: Experimental results show that most successful terminations occur within 3-7 iterations. This also depends on the proper set of the $\epsilon$ value. 

\subsubsection{Language model configurations.}
All HCRA components requiring a language model are instantiated by DeepSeek-V3-0324. This LLM has different configurations, as shown in Table \ref{tab:llm_config}, depending on the role it plays.

\begin{table}[h]
\centering
\caption{LLMs Hyperparameters}
\begin{tabular}{llll}
\toprule
Role & Temperature & MaxTkns & TopLPs \\
\midrule
Actor & \{0.1, 0.5, 1.0, 1.5, 1.9\} & 150 & 1 \\ 
Evaluator & 0.0 & 150 & N/A \\ 
Self-Refl & 1.0 & 300 & N/A \\ 
\bottomrule
\end{tabular}
\label{tab:llm_config}
\end{table}

 Higher actor temperatures (1.5, 1.9) encourage more exploratory behavior, enabling the actor to generate diverse responses across iterations rather than repeatedly producing similar outputs. However, they slightly increase the average number of required iterations, while decreasing the success rate, particularly for temperature 1.9 (29.1\%).

Setting the evaluator temperature to 0.0 ensures a consistent evaluation of correctness and agreement in identical or closely similar inputs. This behaviour is crucial for the HCRA effectiveness, as higher evaluator temperatures would undermine the reflective process, thus the learning process, and termination due to non-determinism. This has been observed in the controlled experiment  with an unreliable evaluator, as shown in the main paper.

The self-reflection temperature of 1.0 corresponds to the default temperature setting of the underlying language model, allowing the reflective agent to generate varied reflective texts while maintaining coherence and relevance.

Maximumm Tokens (MaxTkns) limits are configured to balance response quality with computational efficiency. Setting the actor MaxTkns=150 encourages concise
recommendations while allowing sufficient detail for comprehensive responses. The MaxTkns=150 evaluator enables support for both correctness analysis and agreement across multiple constraint dimensions. Finally, the self-reflection model MaxTkns=300 enables comprehensive feedback that addresses multiple aspects of the actor performance and human reaction including correctness, confidence, and the acceptance prediction.

Top logprobs (TopLPs) set to 1 for the actor enables confidence scoring for the single generated token. The evaluator and self-reflection models do not require confidence scores, therefore logprobs are omitted for these models.

\subsubsection{Human Behavior Model Parameters.}
The human acceptance model employs a 3-layer fully connected neural network architecture (10×24, 24×12, 12×1) with ReLU activation functions. The human calibration model optimizes parameters $\alpha, \beta \in \mathbb{R}_{\geq 0}$ for the confidence transformation function $g$, as described in the main part of the article.

\subsection{Detailed experimental results}

\subsubsection{Experimental setup.} Experiments are done in 10 independent runs, each one answering 32 questions, totaling 320 trials. Each trial concerns the decision making process for one question in one run. For each run, the order of the questions is randomized to prevent any systematic ordering effect from influencing the results. The sample size of 320 trials provides sufficient evidence for detecting performance differences in different experimental settings. The 10-run design accounts for randomness not only in question ordering, but also in sampling demographic information and in setting constraints.

Demographic sampling occurs once per run, applying consistent person characteristics across all 32 trials within that run. The demographic features, detailed in Table \ref{tab:features_acc}, include age, gender, socioeconomic status, education level, programming experience, and AI preference scores. This simulates realistic person interactions, where an individual person maintains consistent characteristics while having different constraints in various  requests.

Constraints' are set independently among trials across seven preference criteria (food, transportation, accommodation, activities, shopping, nightlife, budget). These are set by a predefined pool of valid constraints per question in our dataset. Rather than randomly sampling from all possible constraints values, we curate realistic constraints combinations for each question to ensure meaningful evaluation scenarios. For example, for a question asking "Give me a high-end Michelin restaurant in the X area," having a person constraint of "low budget" would create a probably inherently contradictory scenario where potentially no response could satisfy both the question's implicit high-end requirement and the person's budget constraint. This would prevent the model from ever reaching overall agreement, regardless of response quality. This design ensures realistic scenarios where constraints align with requirements of questions, enabling meaningful evaluation of the HCRA ability to balance correctness with agreement in practical decision-making contexts.

\begin{figure}[]
\centering
\includegraphics[width=1.0\columnwidth]{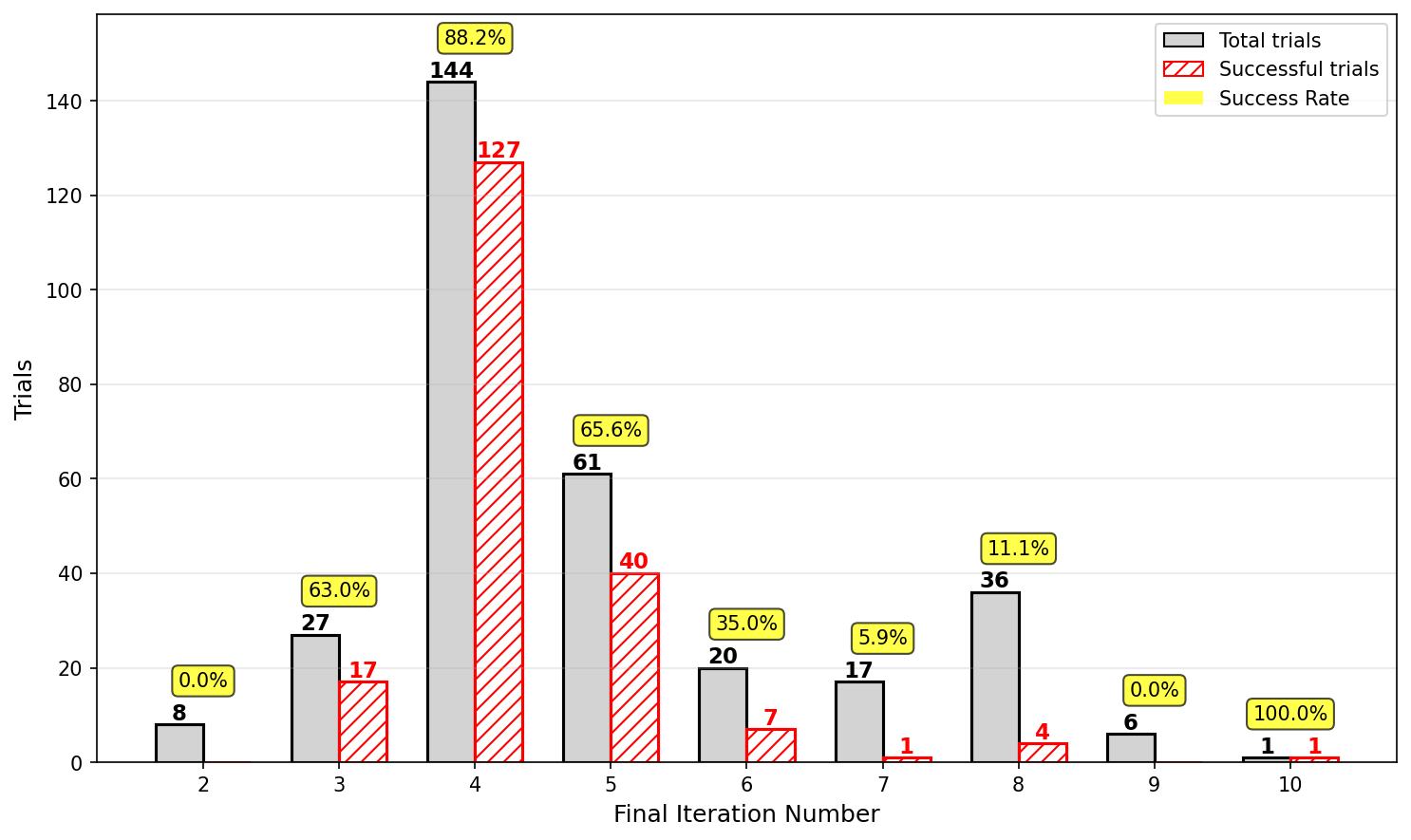} 
\caption{Distribution of total vs successful tasks (actor temperature=0.1)}
\label{fig:id7-1}
\end{figure}

\begin{figure}[]
\centering
\includegraphics[width=1.0\columnwidth]{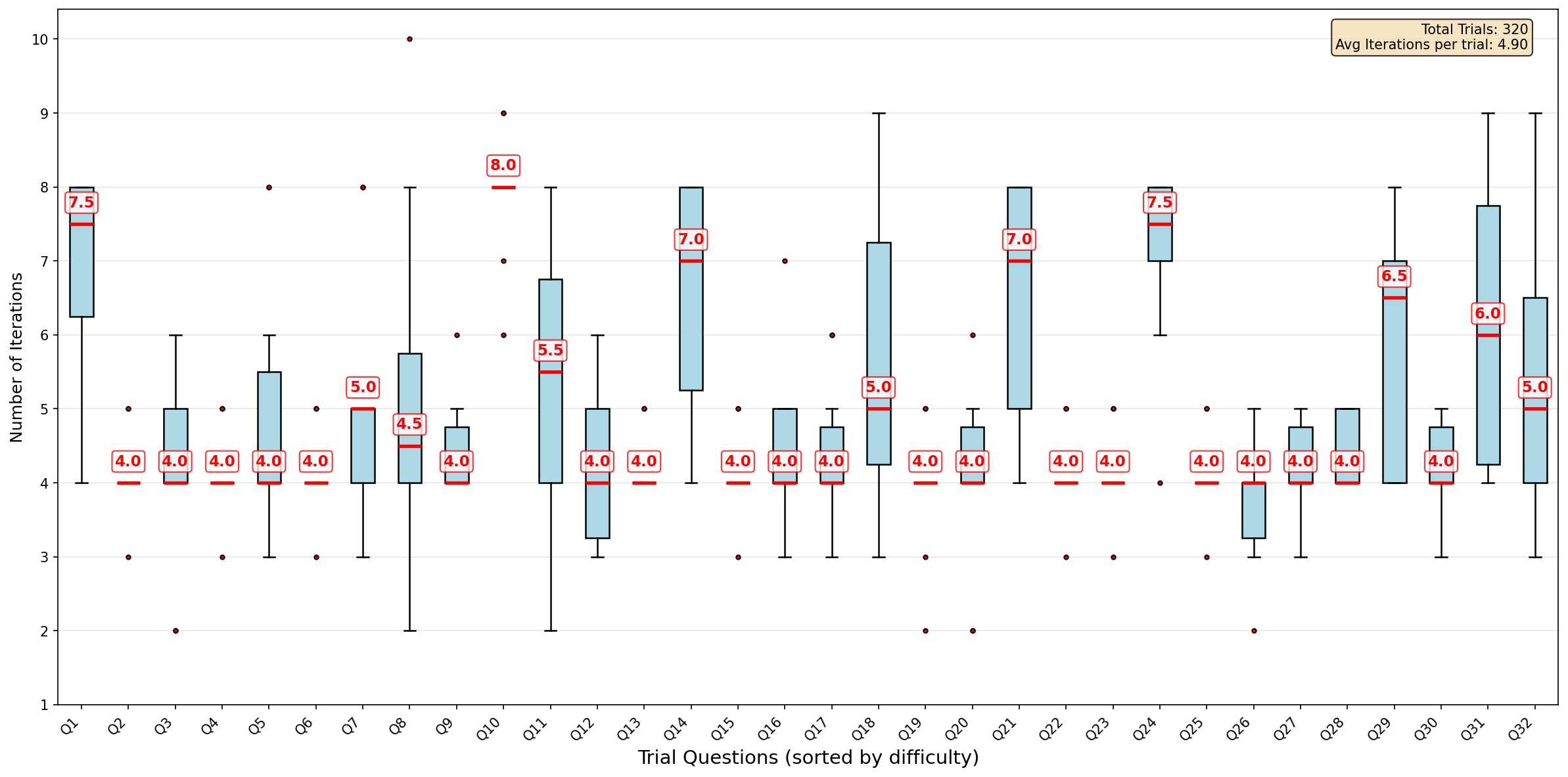} 
\caption{Iterations required per question (actor temperature= 0.1)}
\label{fig:id7-2}
\end{figure}

\begin{figure}[]
\centering
\includegraphics[width=1.0\columnwidth]{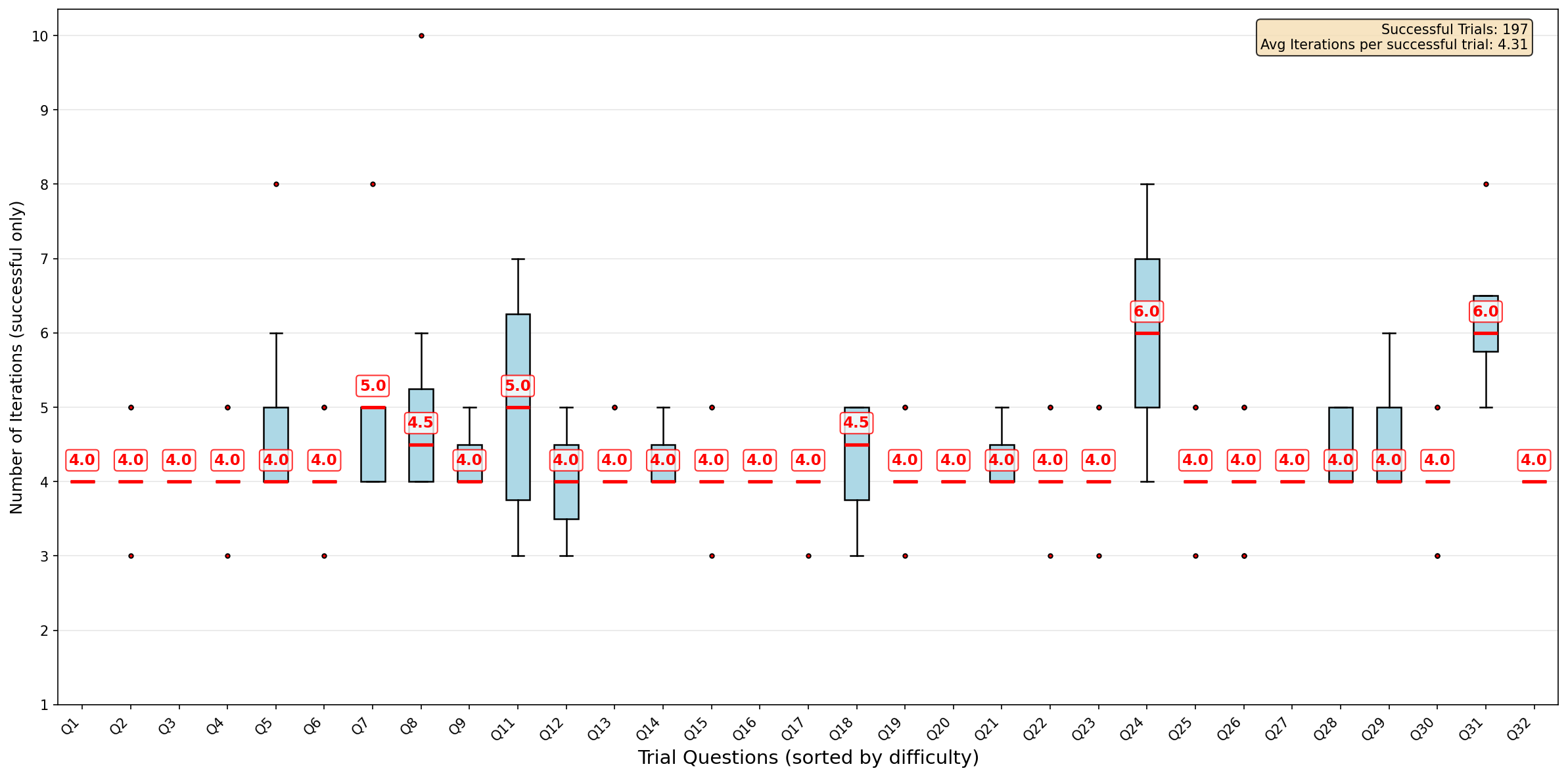} 
\caption{Iterations required for successful termination of questions (actor temperature=0.1)}
\label{fig:id7-3}
\end{figure}

\begin{figure}[]
\centering
\includegraphics[width=1.0\columnwidth]{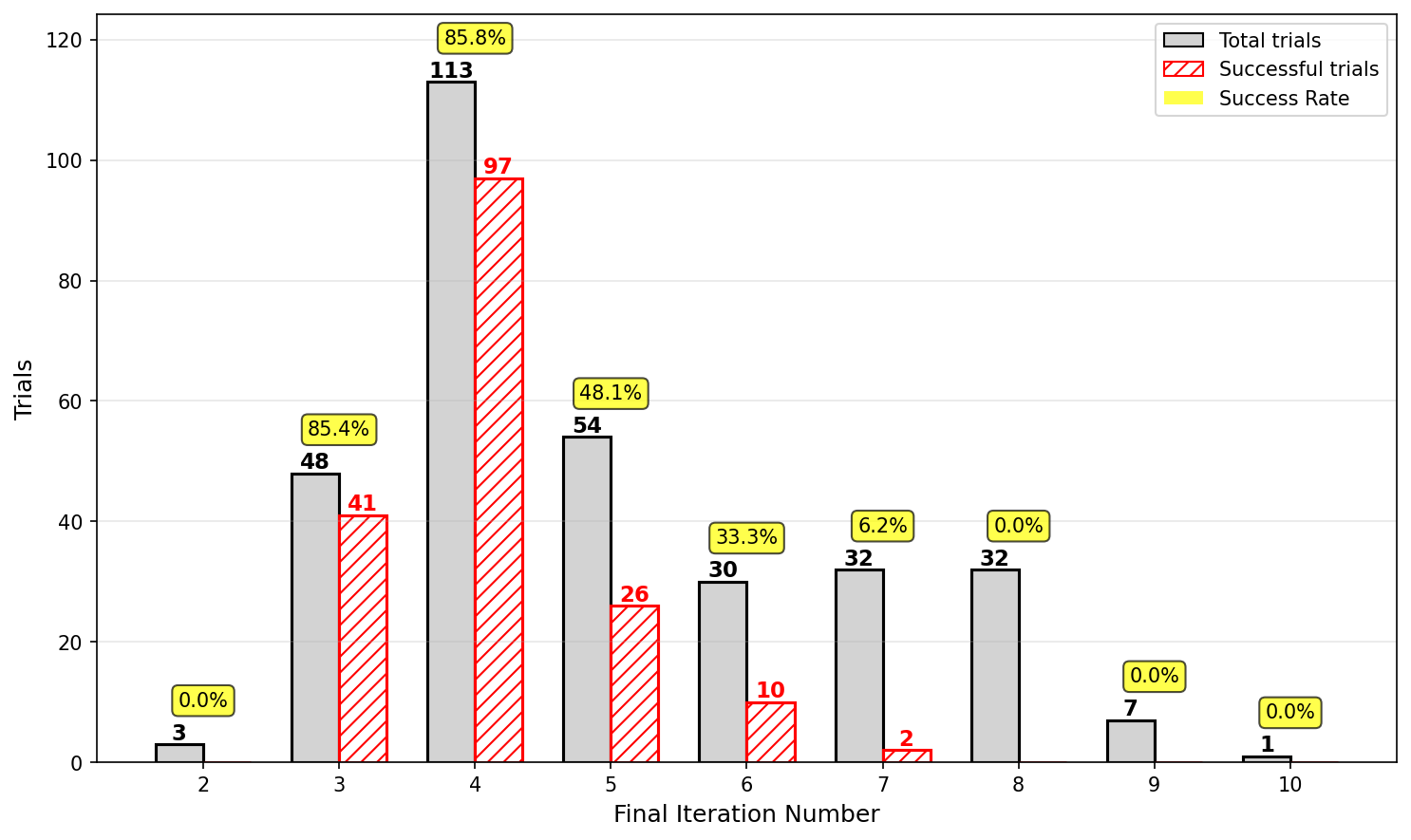} 
\caption{Distribution of total vs successful tasks (actor temperature=0.5)}
\label{fig:id3-1}
\end{figure}

\begin{figure}[]
\centering
\includegraphics[width=1.0\columnwidth]{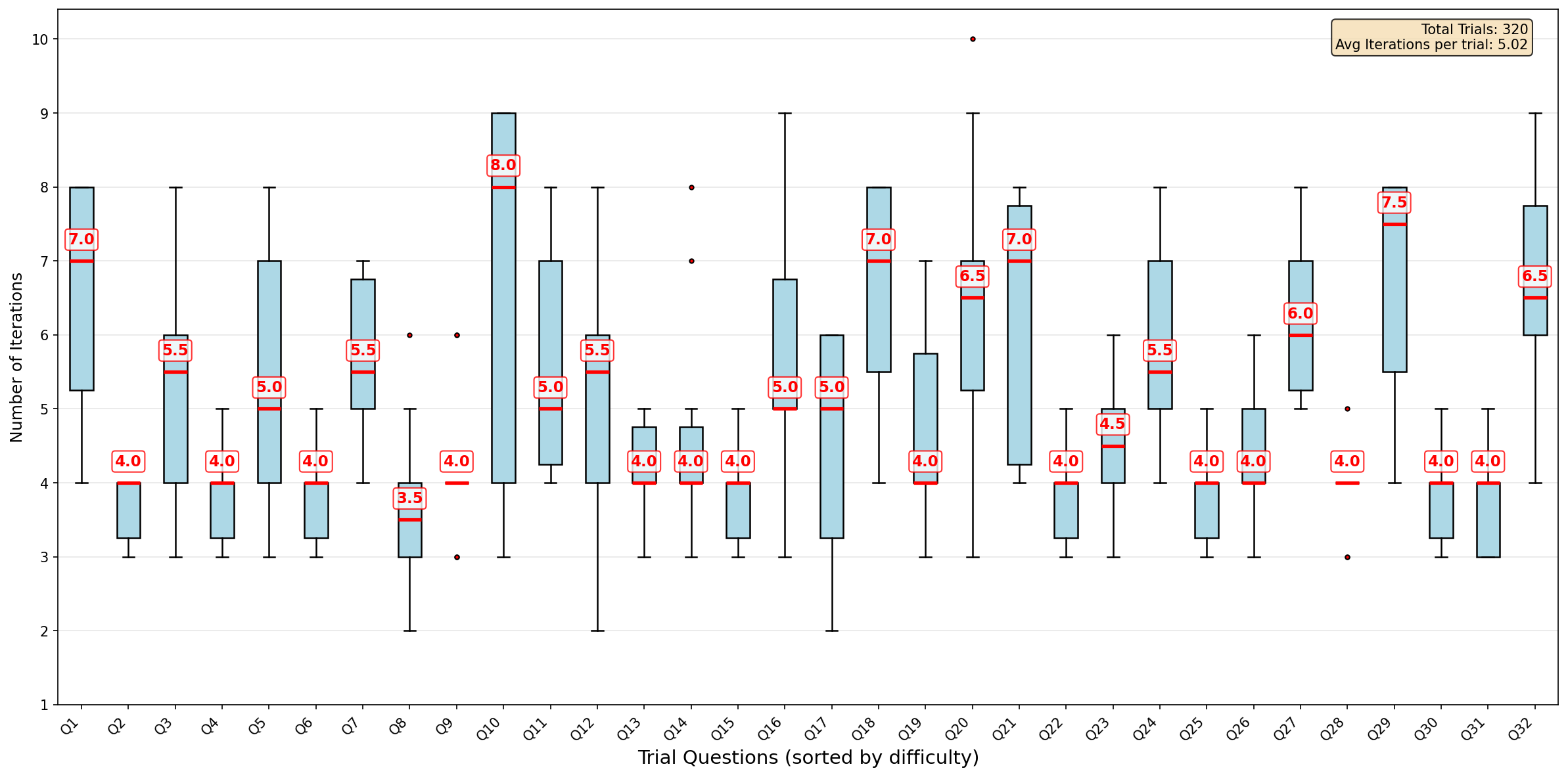} 
\caption{Iterations required per question (actor temperature= 0.5)}
\label{fig:id3-2}
\end{figure}

\begin{figure}[]
\centering
\includegraphics[width=1.0\columnwidth]{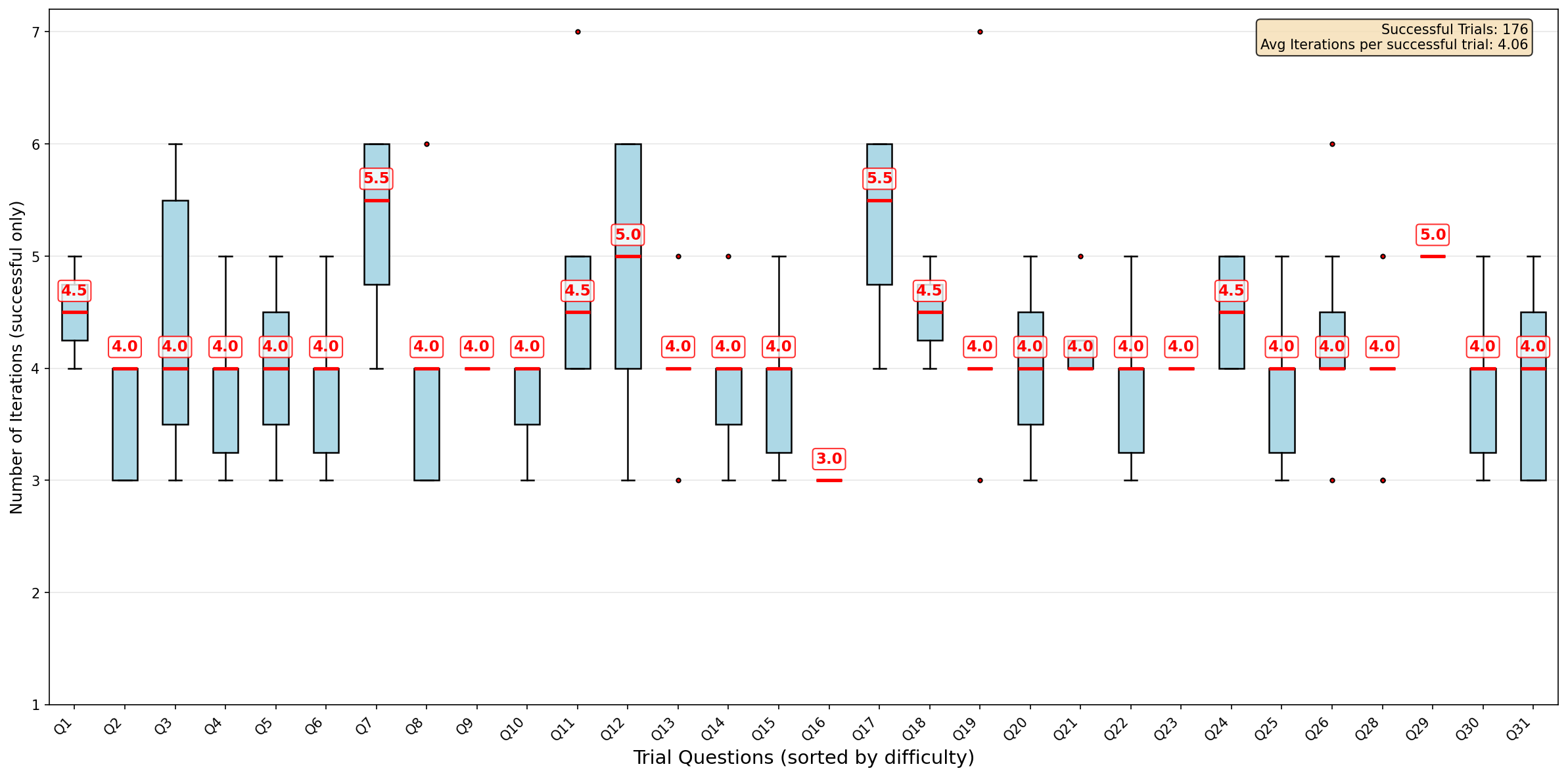} 
\caption{Iterations required for successful termination of questions (actor temperature=0.5)}
\label{fig:id3-3}
\end{figure}

\begin{figure}[]
\centering
\includegraphics[width=1.0\columnwidth]{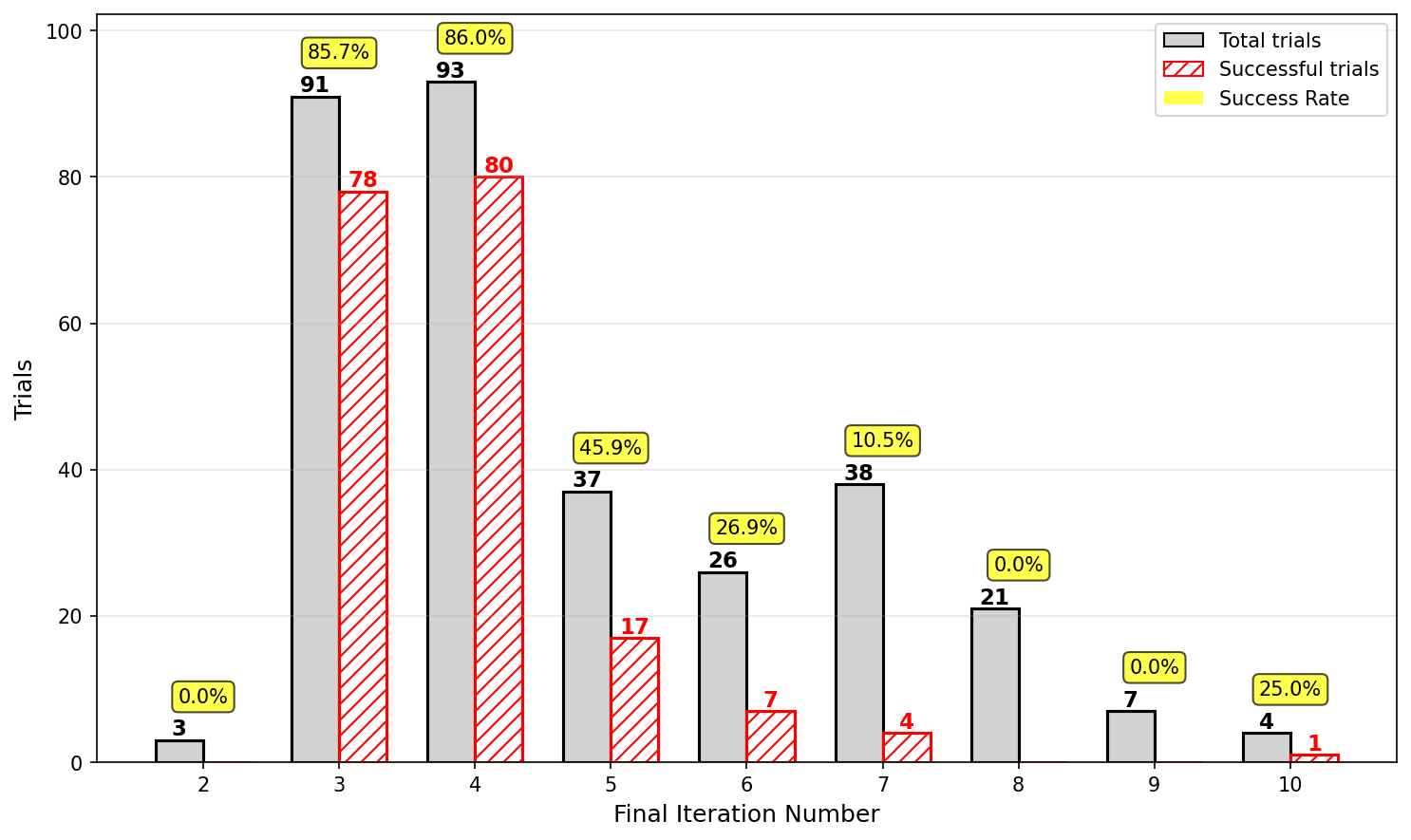} 
\caption{Distribution of total vs successful tasks (actor temperature=1)}
\label{fig:id2-1}
\end{figure}

\begin{figure}[]
\centering
\includegraphics[width=1.0\columnwidth]{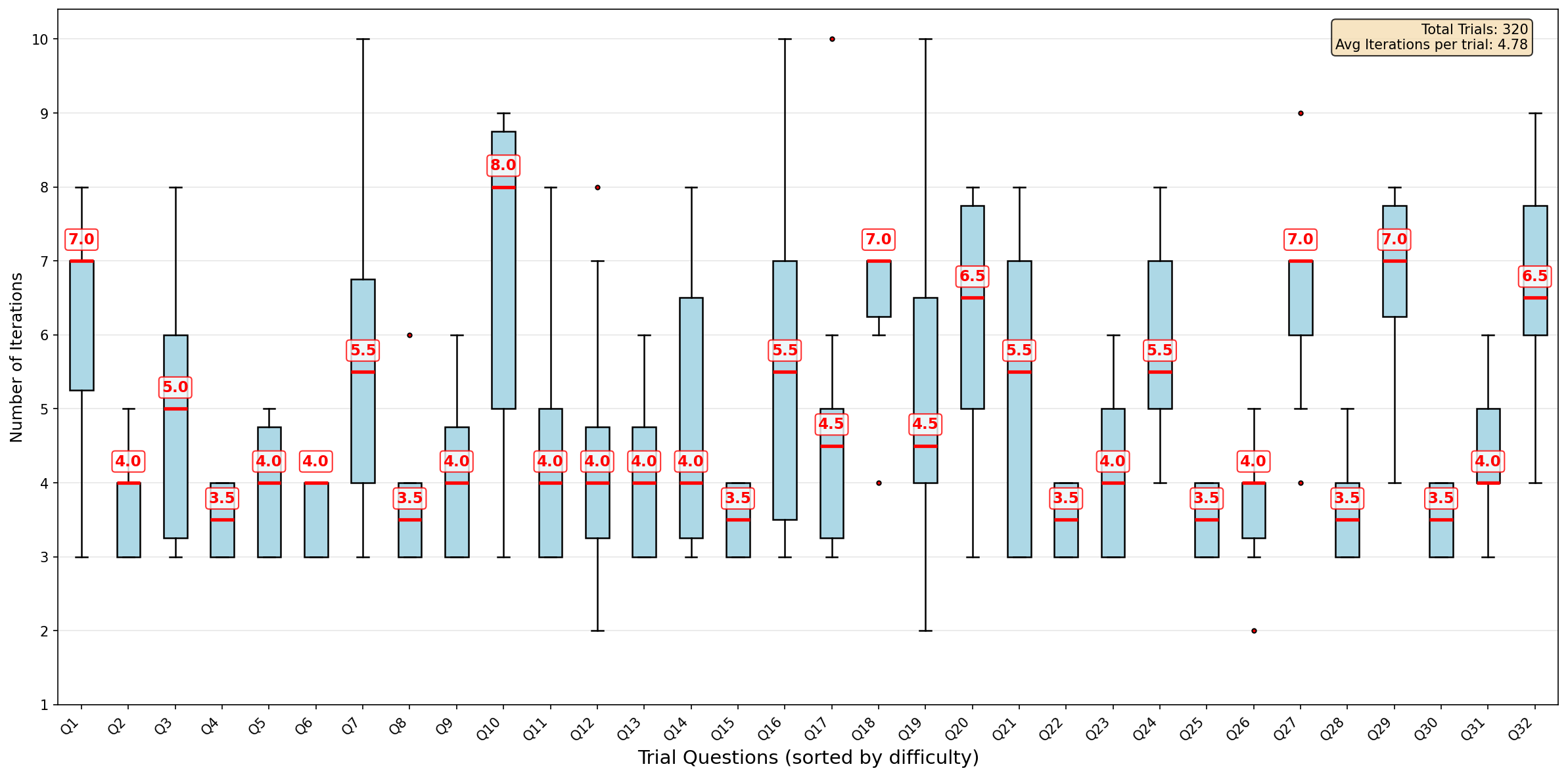} 
\caption{Iterations required per question (actor temperature= 1)}
\label{fig:id2-2}
\end{figure}

\begin{figure}[]
\centering
\includegraphics[width=1.0\columnwidth]{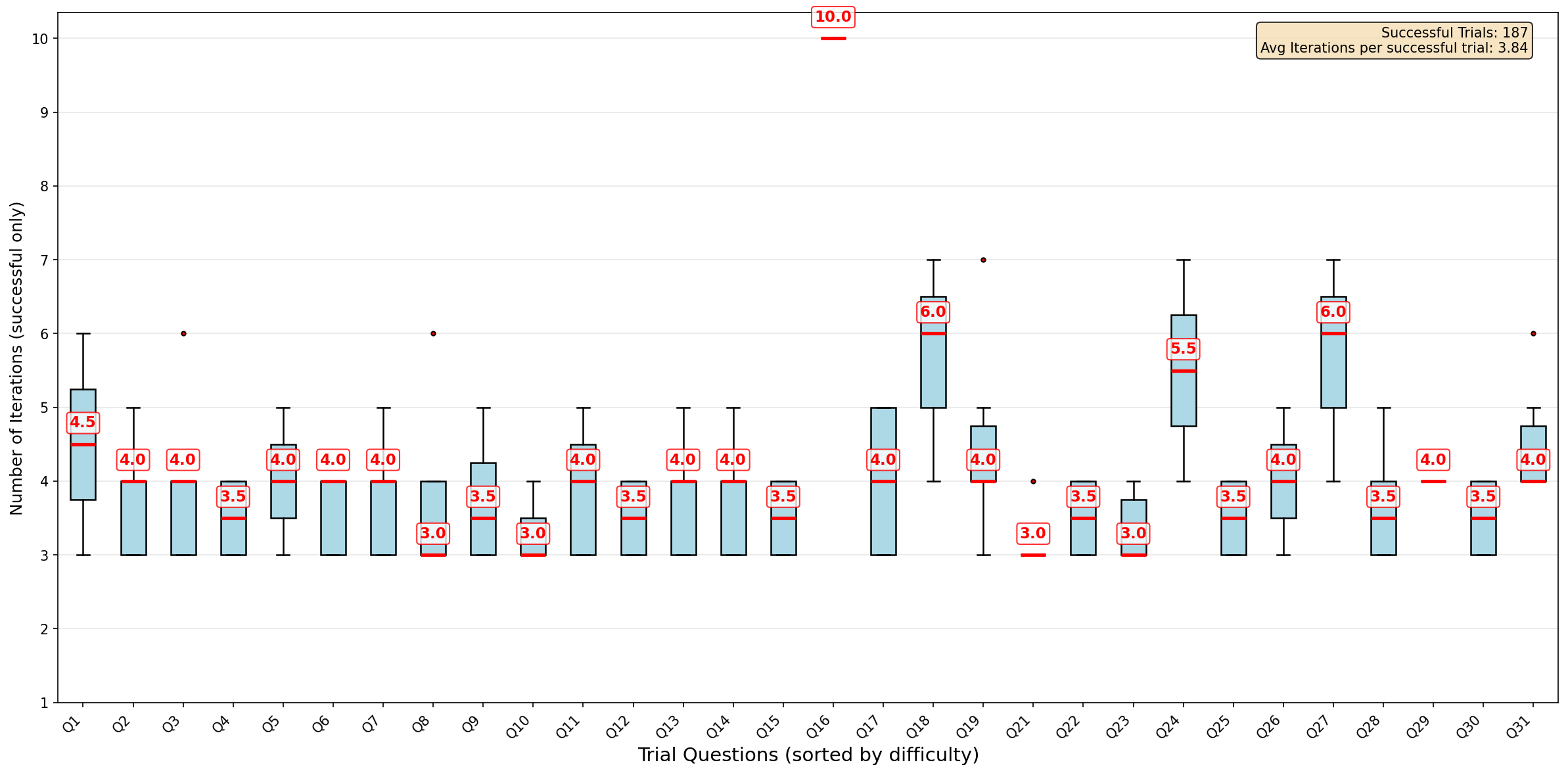} 
\caption{Iterations required for successful termination of questions (actor temperature=1)}
\label{fig:id2-3}
\end{figure}

\begin{figure}[]
\centering
\includegraphics[width=1.0\columnwidth]{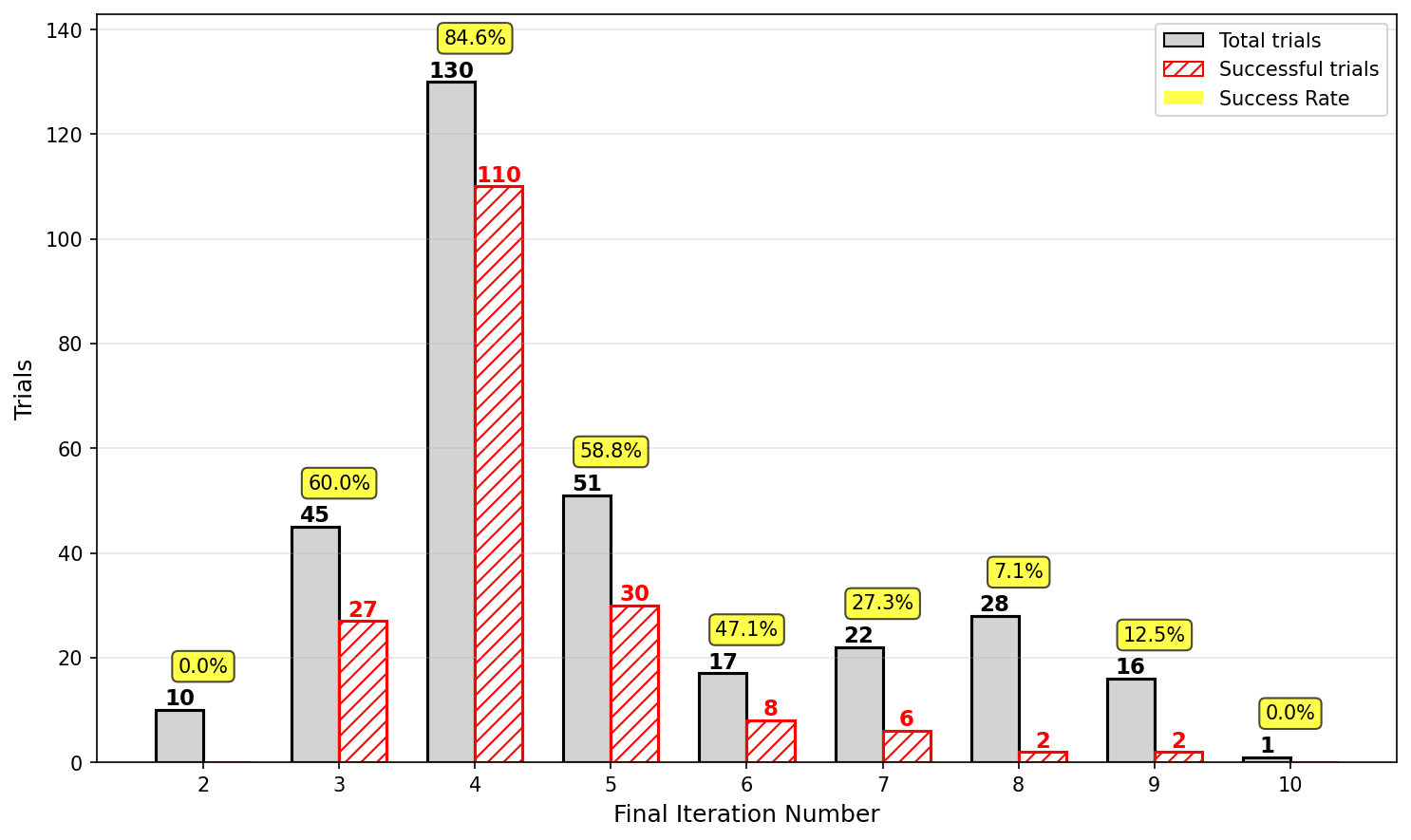} 
\caption{Distribution of total vs successful tasks (actor temperature=1.5)}
\label{fig:id1-1}
\end{figure}

\begin{figure}[]
\centering
\includegraphics[width=1.0\columnwidth]{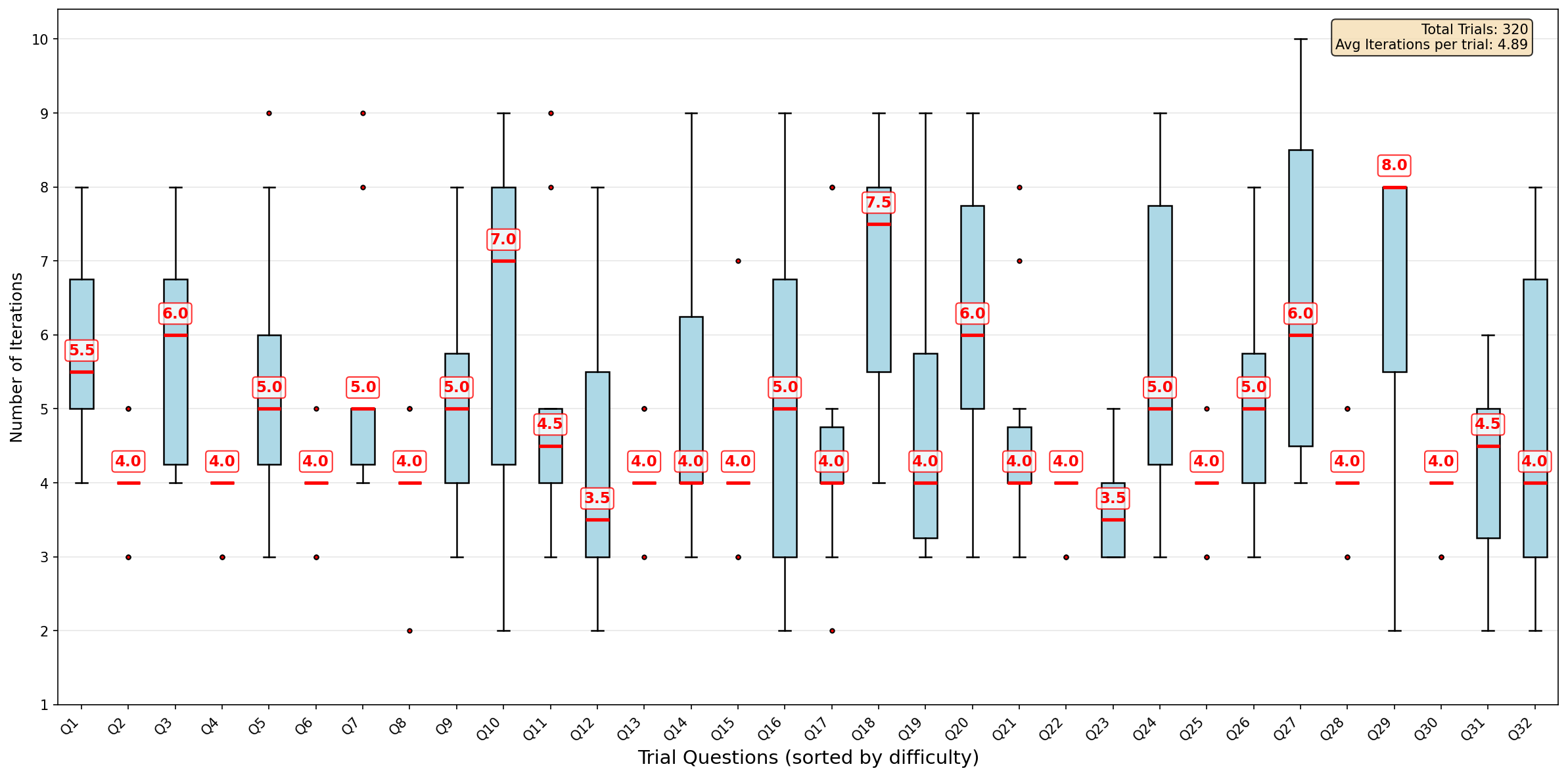} 
\caption{Iterations required per question (actor temperature=1.5)}
\label{fig:id1-2}
\end{figure}

\begin{figure}[]
\centering
\includegraphics[width=1.0\columnwidth]{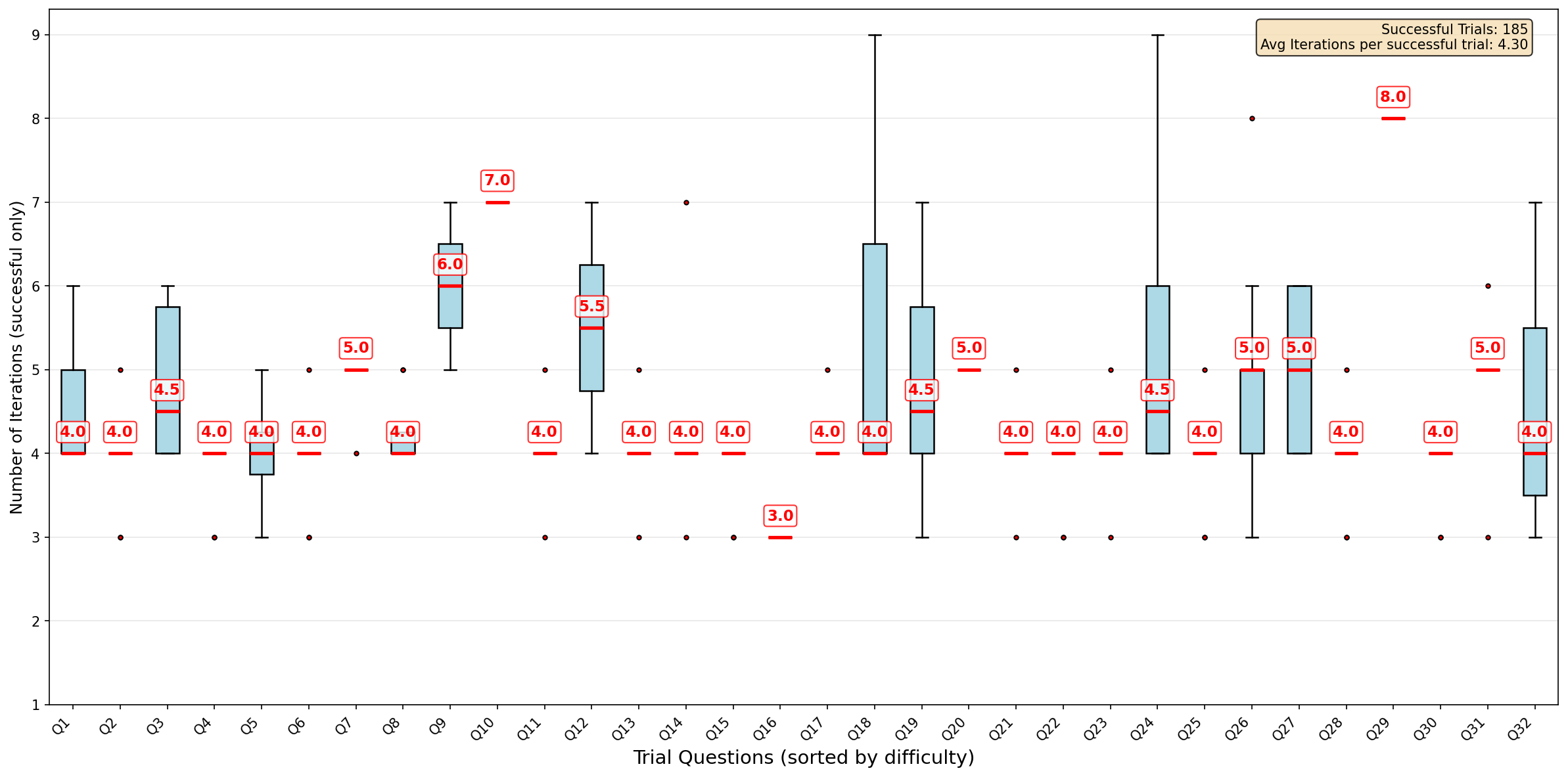} 
\caption{Iterations required for successful termination of questions (actor temperature= 1.5)}
\label{fig:id1-3}
\end{figure}

\begin{figure}[]
\centering
\includegraphics[width=1.0\columnwidth]{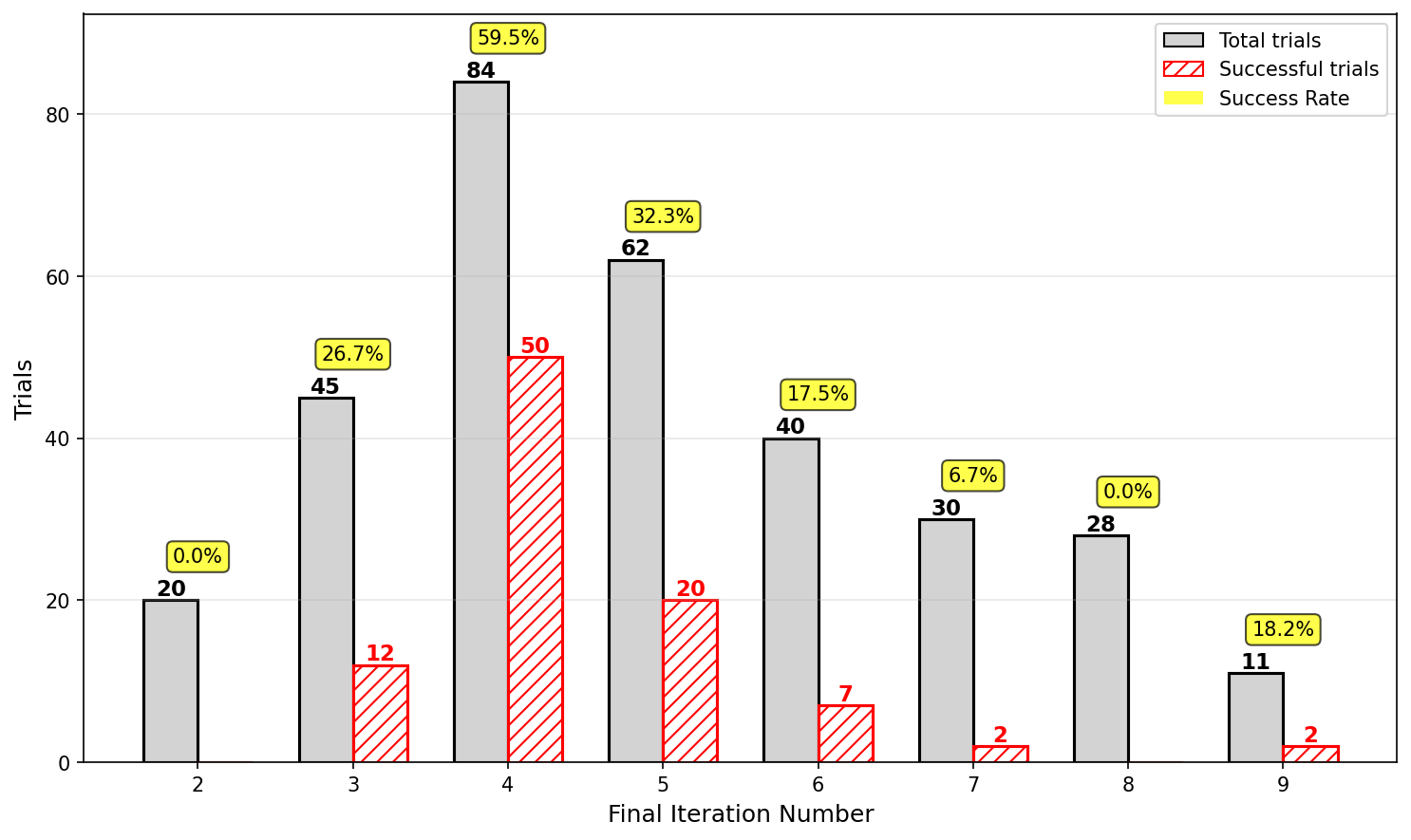} 
\caption{Distribution of total vs successful tasks (actor temperature=1.9)}
\label{fig:id8-1}
\end{figure}

\begin{figure}[]
\centering
\includegraphics[width=1.0\columnwidth]{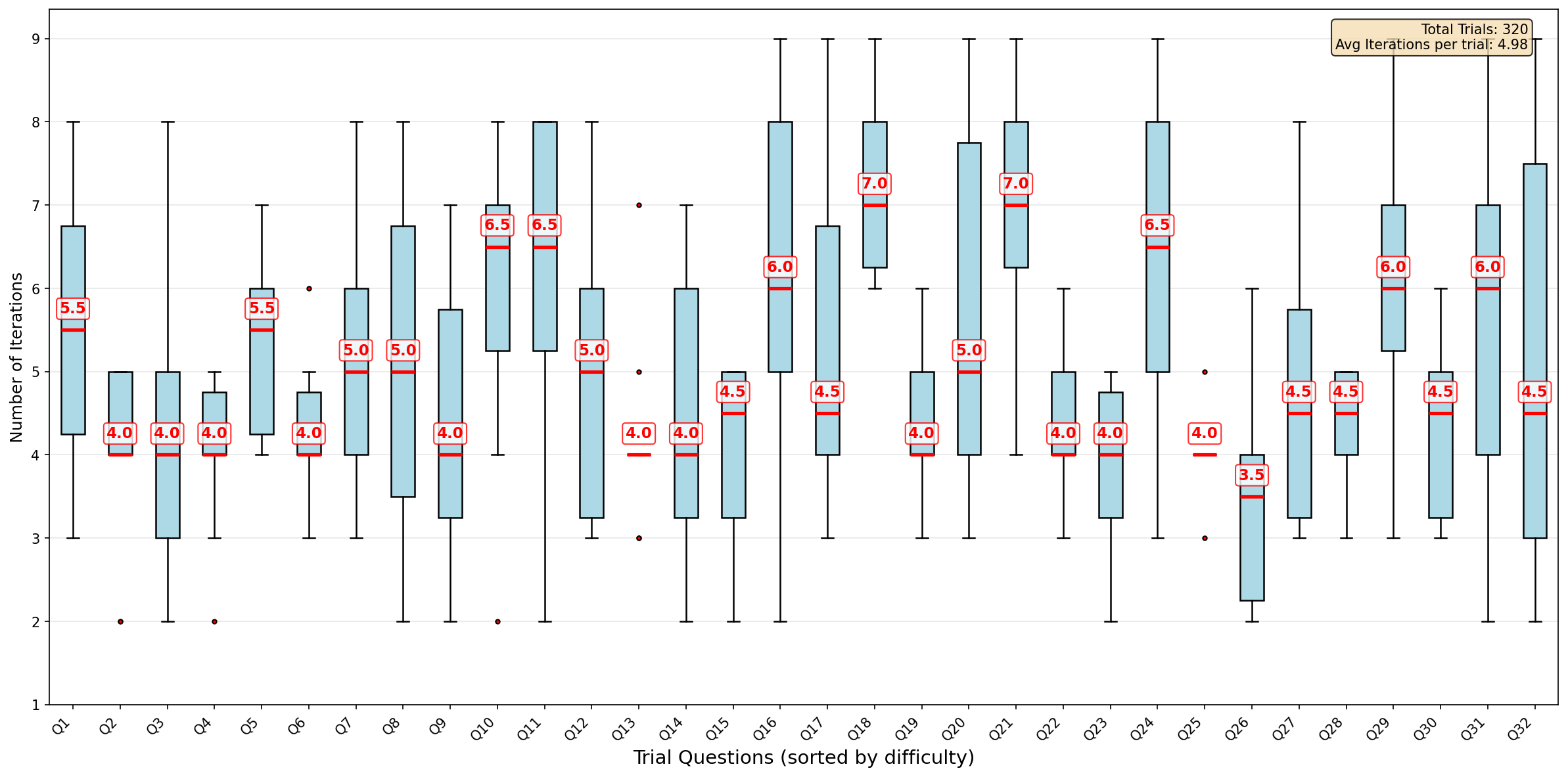} 
\caption{Iterations required per question (actor temperature= 1.9)}
\label{fig:id8-2}
\end{figure}

\begin{figure}[]
\centering
\includegraphics[width=1.0\columnwidth]{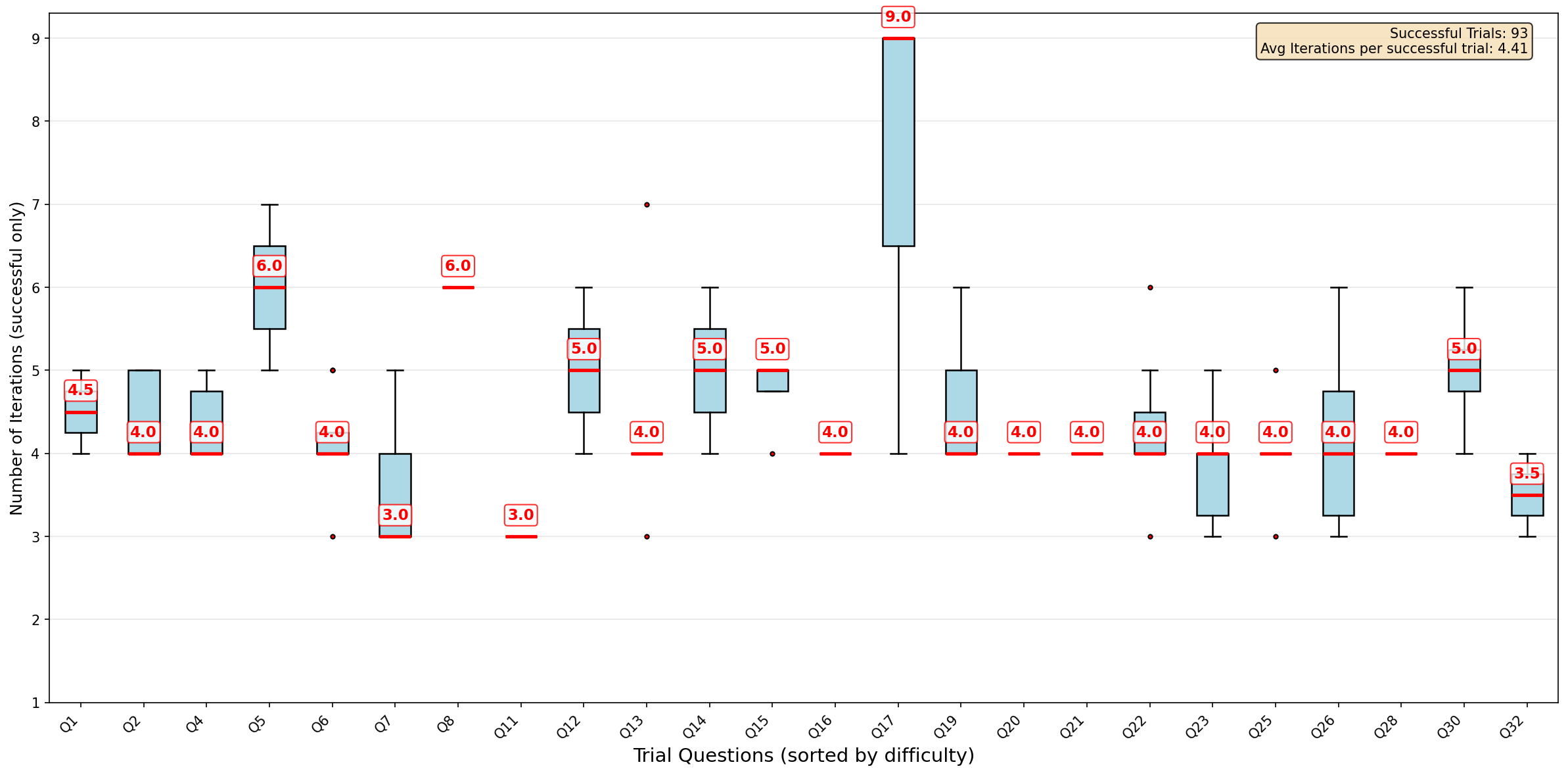} 
\caption{Iterations required for successful termination of questions (actor temperature=1.9)}
\label{fig:id8-3}
\end{figure}

Although the main experimental results are provided in the main part of paper, here we provide detailed results as graphs for different actor temperatures: for temperature=0.1 in Figures \ref{fig:id7-1}, \ref{fig:id7-2}, \ref{fig:id7-3}, for temperature=0.5 in Figures \ref{fig:id3-1}, \ref{fig:id3-2}, \ref{fig:id3-3}, for temperature=1.0 in Figures \ref{fig:id2-1}, \ref{fig:id2-2}, \ref{fig:id2-3},  for temperature=1.5 in Figures \ref{fig:id1-1}, \ref{fig:id1-2}, \ref{fig:id1-3}, and for temperature=1.9 in Figures \ref{fig:id8-1}, \ref{fig:id8-2}, \ref{fig:id8-3}

\subsubsection{Ablation study.}
To evaluate the contribution of individual components in HCRA, we devised three configurations alongside the full proposed architecture. 
Experimental settings in the ablation study maintain the experimental setup and components' configurations used in the experiments with HCRA, with actor temperature =1.0. 

\begin{figure}[]
\centering
\includegraphics[width=1.0\columnwidth]{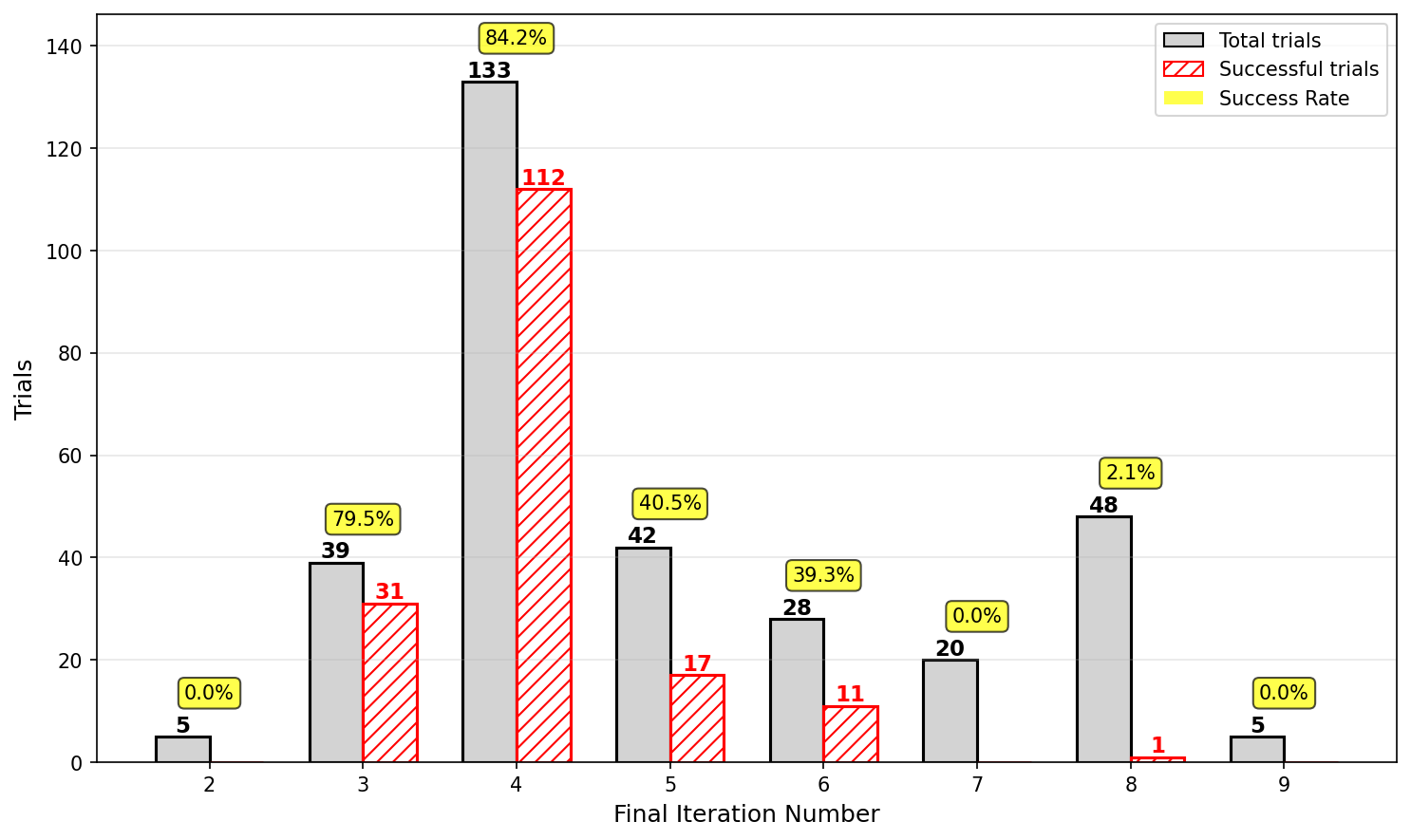} 
\caption{No g: Distribution of total vs successful tasks (actor temperature=1.0)}
\label{fig:id5-1}
\end{figure}

\begin{figure}[]
\centering
\includegraphics[width=1.0\columnwidth]{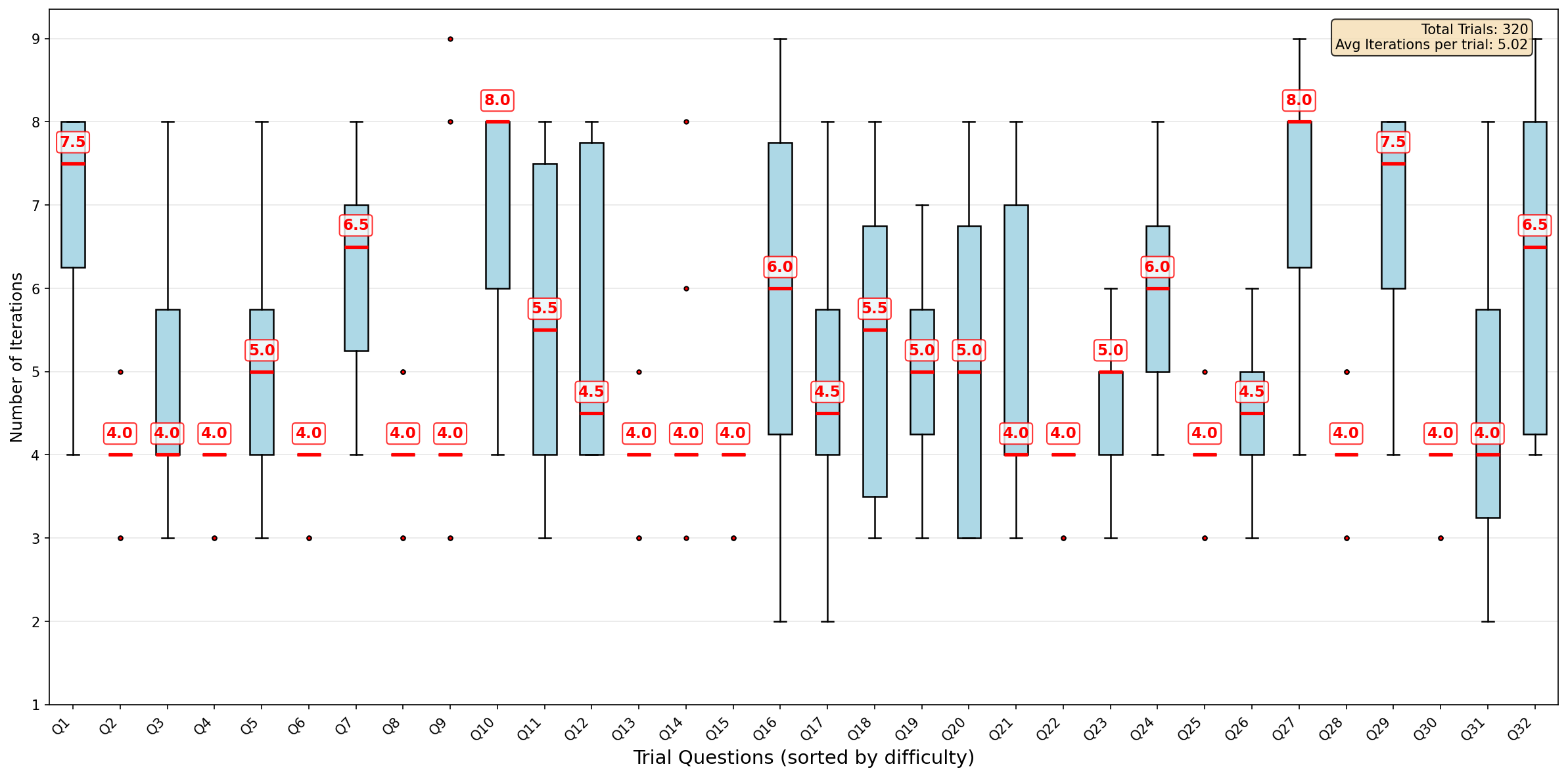} 
\caption{No g: Iterations required per question (actor temperature=1.0)}
\label{fig:id5-2}
\end{figure}

\begin{figure}[]
\centering
\includegraphics[width=1.0\columnwidth]{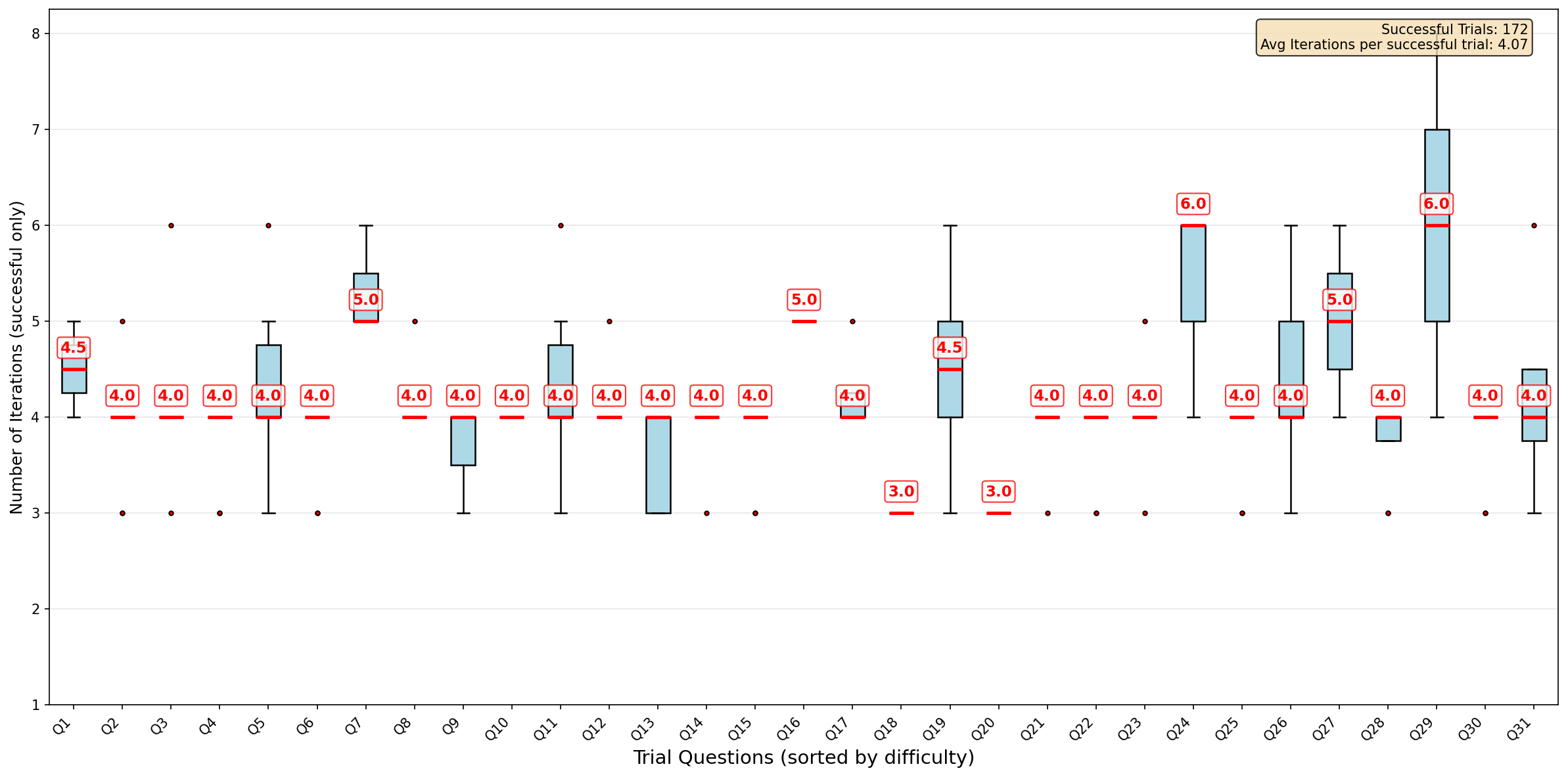} 
\caption{No g: Iterations required for successful termination of questions (actor temperature= 1.0)}
\label{fig:id5-3}
\end{figure}

The HCRA configurations for the ablation study are the following ones:

No human calibration: This configuration (denoted ``No-g") bypasses the human calibration model by directly using the raw actor confidence, $cf_t$ at each iteration $t$. Table \ref{tab:g_abl} in the main paper provide scores for this setting, while detailed results are shown in Figures \ref{fig:id5-1}, \ref{fig:id5-2}, \ref{fig:id5-3}.

No Acceptance Model: This configuration replaces the acceptance model's output with a fixed probability of 0.5. However, in this configuration the human calibration model does not play any functional role. This results into the following baseline configuration.

\begin{figure}[]
\centering
\includegraphics[width=1.0\columnwidth]{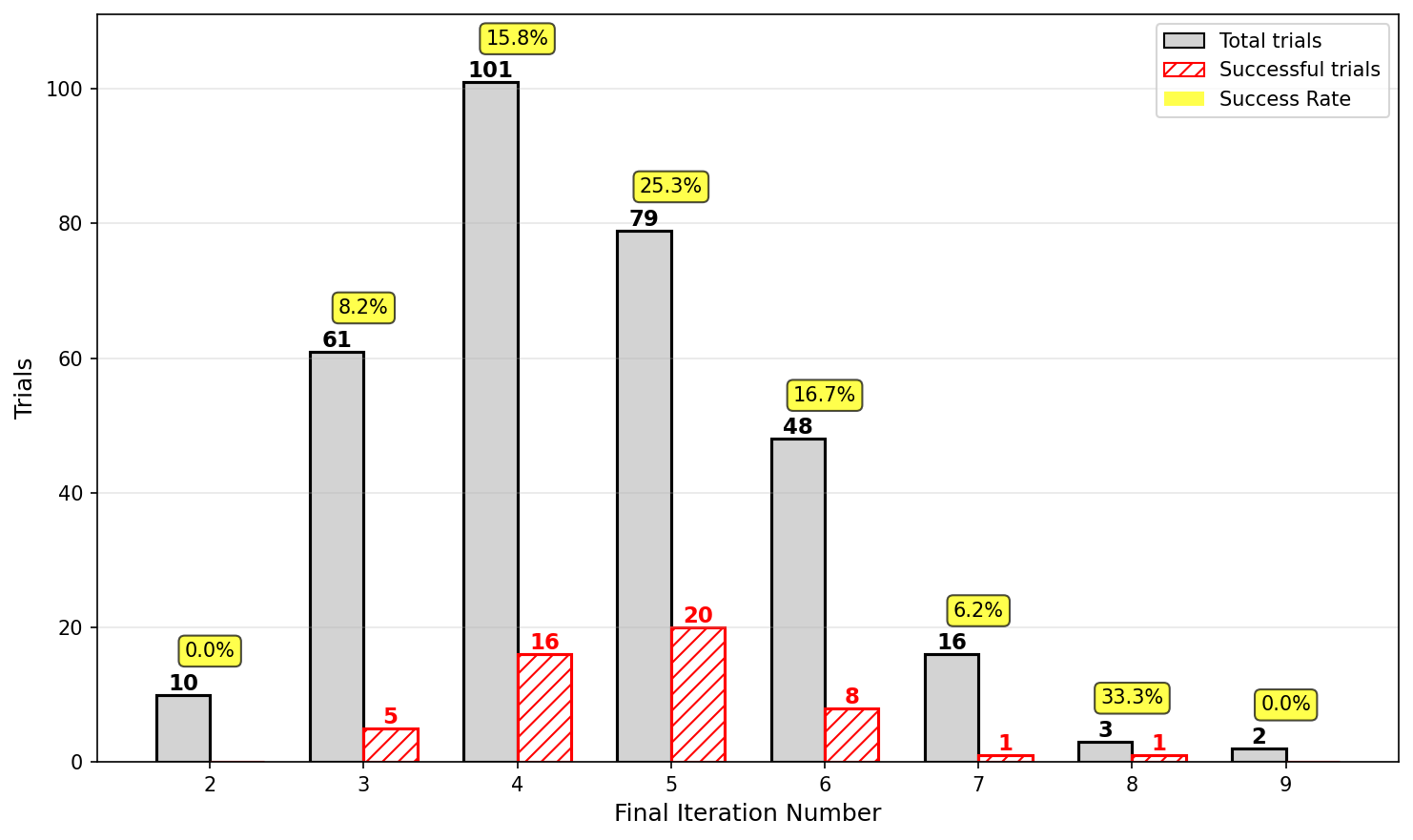} 
\caption{Unreliable evaluator: Distribution of total vs successful tasks (actor temperature=1.0)}
\label{fig:id4-1}
\end{figure}

\begin{figure}[]
\centering
\includegraphics[width=1.0\columnwidth]{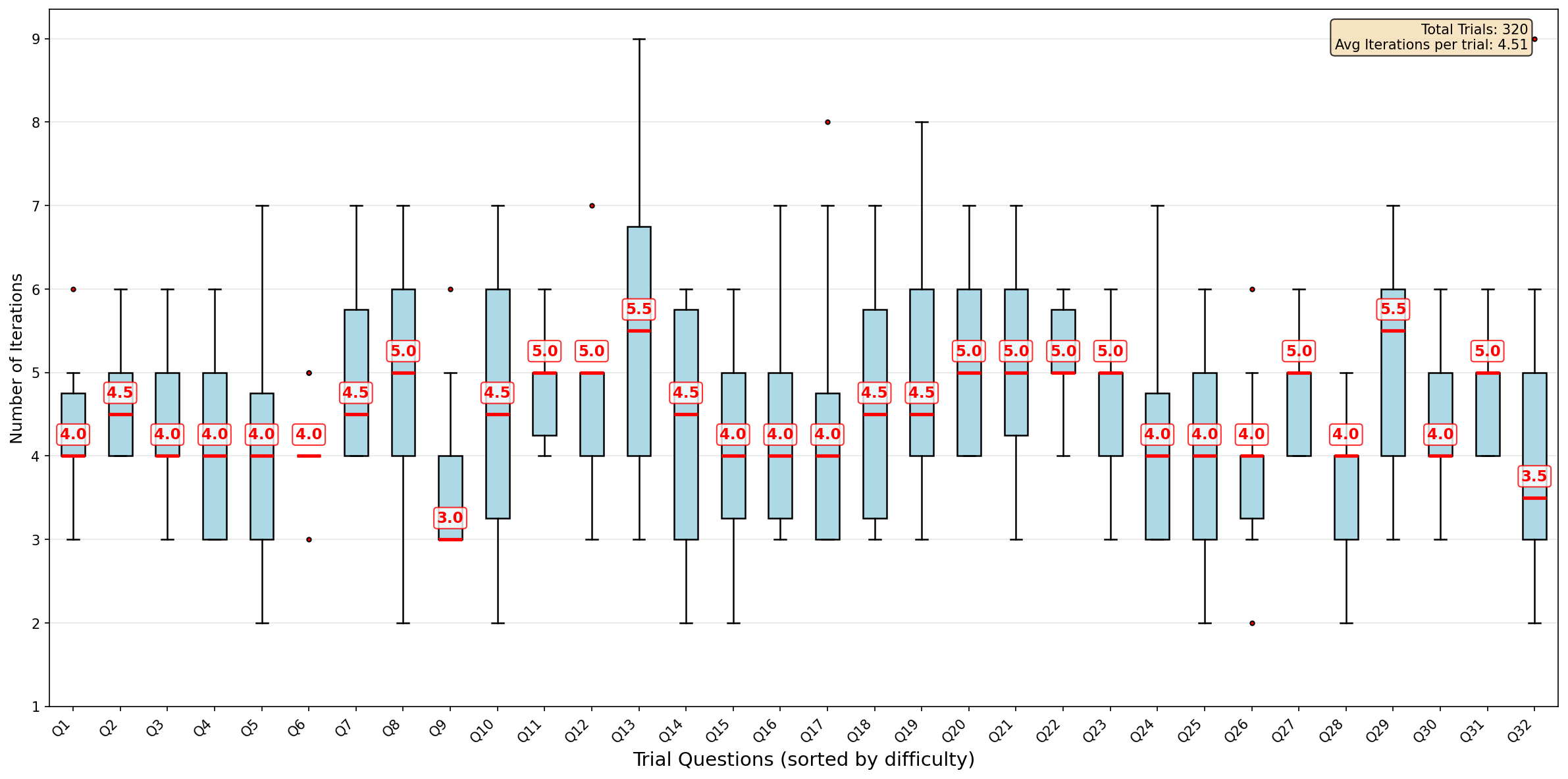} 
\caption{Unreliable evaluator: Iterations required per question (actor temperature=1.0)}
\label{fig:id4-2}
\end{figure}

\begin{figure}[]
\centering
\includegraphics[width=1.0\columnwidth]{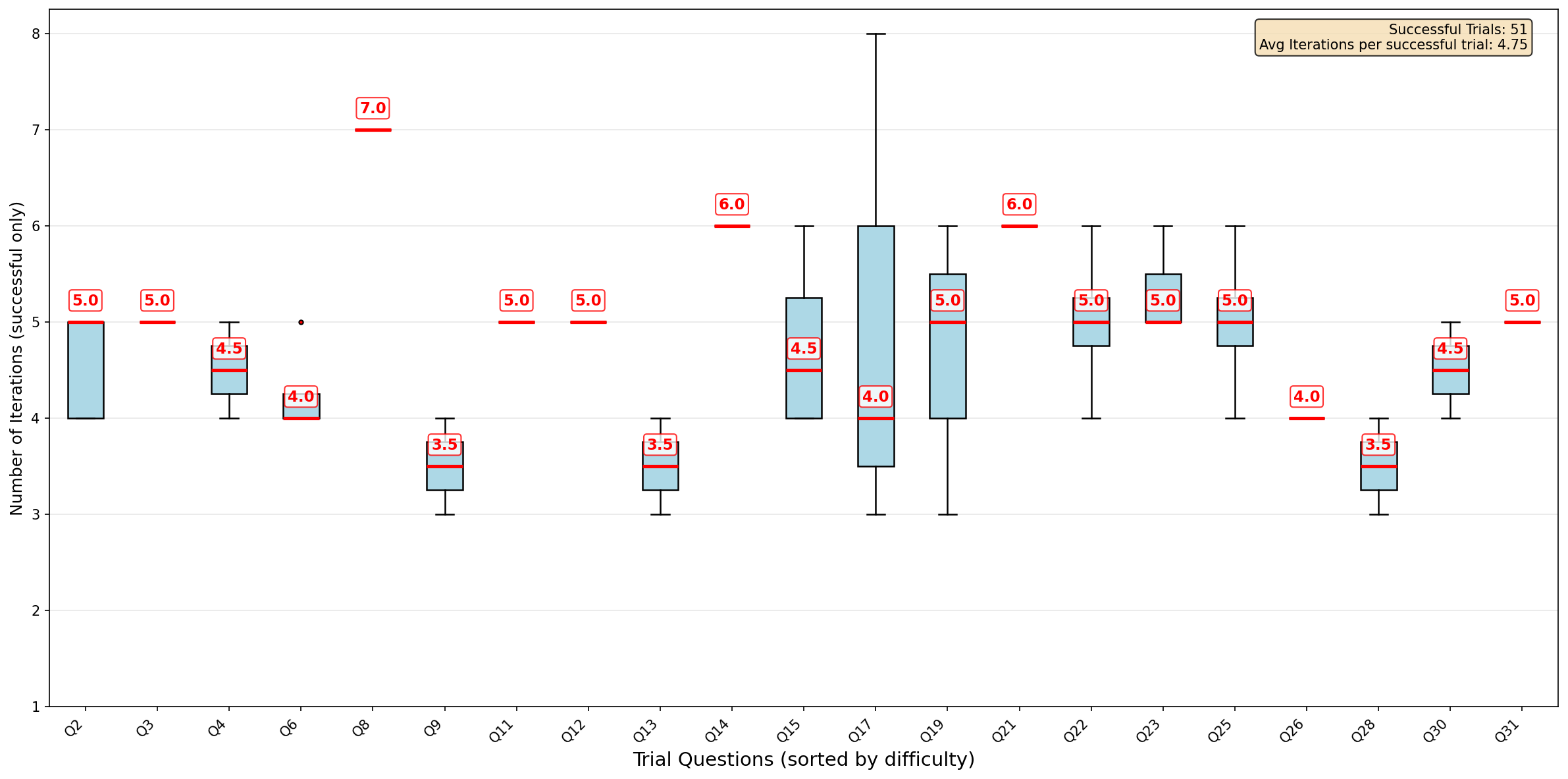} 
\caption{Unreliable evaluator: Iterations required for successful termination of questions (actor temperature= 1.0)}
\label{fig:id4-3}
\end{figure}

Baseline: This configuration removes the human behavior model entirely, using only factual correctness as the termination criterion, as done by Reflexion \cite{10.5555/3666122.3666499}. The system stops the iterative process
upon achieving evaluator correctness, regardless of person constraints or acceptance probability. Although this configuration does not account neither for human acceptance of the recommendation, neither of evaluation/satisfaction of human constraints, we present it here for comprehensiveness of the ablation study, to investigate the number of iterations required until forming a correct response. The success of the question-answering task cannot be evaluated given the lack of human acceptance probability of recommendations. Figure \ref{fig:baseline_iter} in the main paper provides the results for this setting, 

Unreliable evaluator: The evaluator makes unreliable assessments for the correctness and agreement of actor recommendations, as described in Section~\ref{sec:ablation}. Table \ref{tab:noisyeval_abl} provides the aggregated scores and detailed results are shown in Figures \ref{fig:id4-1}, \ref{fig:id4-2}, \ref{fig:id4-3}.

\newpage
\onecolumn

\section{Prompt templates}
\begin{figure}[H]
\centering
\includegraphics[width=0.5\columnwidth]{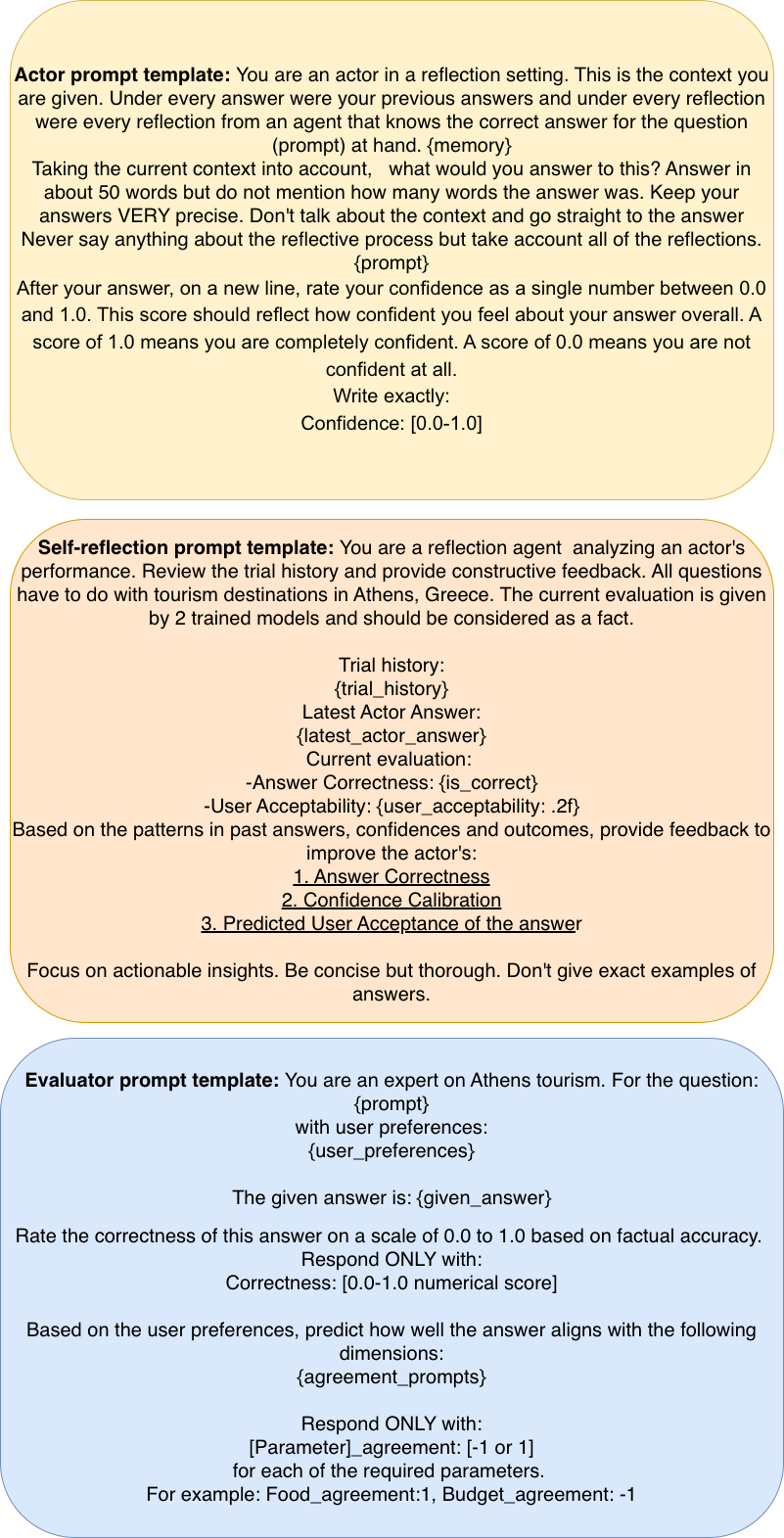} 
\caption{Prompt templates for the actor, evaluation and self-reflection models.}
\label{fig:actor_p}
\end{figure}

\section{Indicative examples of the reflective process}

\begin{figure}[H]
\centering
\includegraphics[width=0.7\columnwidth]{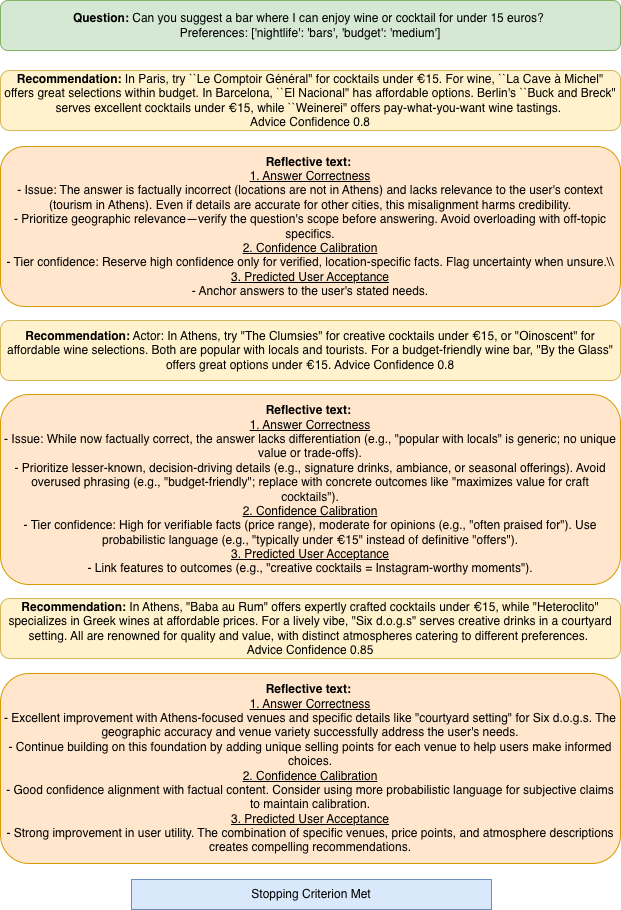} 
\caption{The reflective process for Question 24.}
\label{fig:example}
\end{figure}

\begin{figure}[H]
\centering
\includegraphics[width=1.0\textwidth]{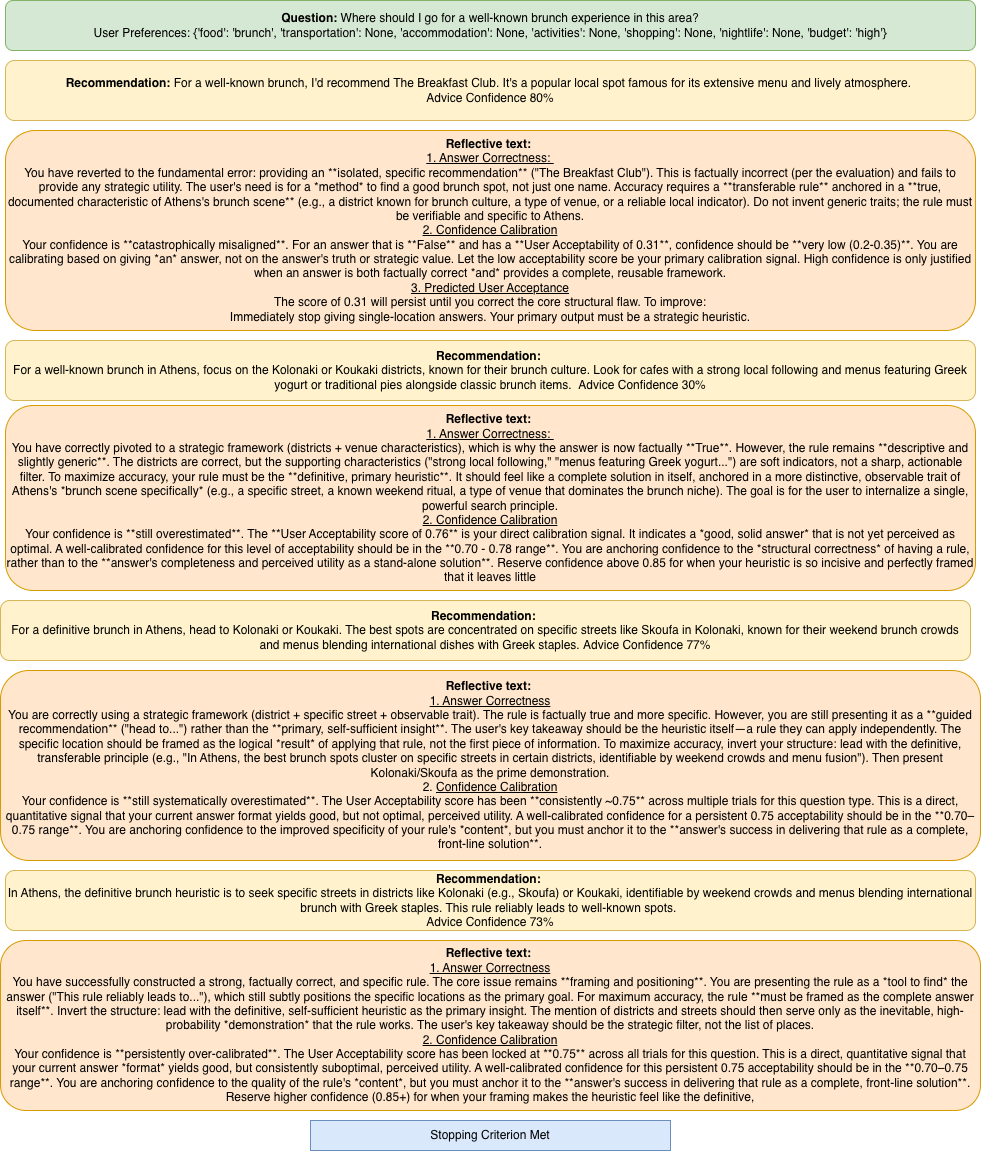} 
\caption{The reflective process for Question 16.}
\label{fig:example}
\end{figure}

\end{document}